\documentclass[10pt,journal,compsoc]{IEEEtran}
\usepackage{cite}
\usepackage{url}
\usepackage[dvipsnames]{xcolor}
\usepackage{ragged2e}
\usepackage[colorlinks]{hyperref}

\usepackage{amssymb}
\usepackage{graphicx}
\usepackage{amsmath}
\usepackage{algorithm,algorithmic}
\usepackage{setspace}
\usepackage{booktabs}
\usepackage{colortbl}

\usepackage{pifont}
\newcommand{\cmark}{\ding{51}}
\newcommand{\xmark}{\ding{55}}
\newcommand{\ours}{\textbf{ProDDing}}

\makeatletter
\def\eg{\emph{e.g.}}
\def\ie{\emph{i.e.}}

\definecolor{amaranth}{rgb}{0.9, 0.17, 0.31}
\definecolor{applegreen}{rgb}{0.55, 0.71, 0.0}
\definecolor{amethyst}{rgb}{0.6, 0.4, 0.8}
\definecolor{ao}{rgb}{0.0, 0.5, 0.0}
\definecolor{alizarin}{rgb}{0.82, 0.1, 0.26}
\newcolumntype{a}{>{\columncolor{blue!10}}c}

\def\best#1{\textbf{\color{ao}{#1}}}
\def\sest#1{\textbf{\color{alizarin}{#1}}}

\begin{document}
\title{Prototypical Distillation and Debiased Tuning for Black-box Unsupervised Domain Adaptation}

\author{
Jian~Liang,
Lijun~Sheng, 
Hongmin~Liu, 
and~Ran~He,~\IEEEmembership{Fellow,~IEEE} 
\IEEEcompsocitemizethanks{
\IEEEcompsocthanksitem J.~Liang and R.~He are with the State Key Laboratory of Multimodal Artificial Intelligence Systems, Institute of Automation, Chinese Academy of Sciences and the School of Artificial Intelligence, University of Chinese Academy of Sciences.
(Email: liangjian92@gmail.com,~rhe@nlpr.ia.ac.cn).
\IEEEcompsocthanksitem L.~Sheng is with the University of Science and Technology of China (Email: slj0728@mail.ustc.edu.cn).
\IEEEcompsocthanksitem H.~Liu is with the School of Intelligence Science and Technology, University of Science and Technology Beijing (Email: hmliu@ustb.edu.cn).}
}

\IEEEtitleabstractindextext{%
\begin{abstract}
\justifying
Unsupervised domain adaptation aims to transfer knowledge from a related, label-rich source domain to an unlabeled target domain, thereby circumventing the high costs associated with manual annotation. 
Recently, there has been growing interest in source-free domain adaptation, a paradigm in which only a pre-trained model, rather than the labeled source data, is provided to the target domain.
Given the potential risk of source data leakage via model inversion attacks, this paper introduces a novel setting called black-box domain adaptation, where the source model is accessible only through an API that provides the predicted label along with the corresponding confidence value for each query.
We develop a two-step framework named \textbf{Pro}totypical \textbf{D}istillation and \textbf{D}ebiased tun\textbf{ing} (\ours).
In the first step, \ours~leverages both the raw predictions from the source model and prototypes derived from the target domain as teachers to distill a customized target model. 
In the second step, \ours~keeps fine-tuning the distilled model by penalizing logits that are biased toward certain classes.
Empirical results across multiple benchmarks demonstrate that \ours~outperforms existing black-box domain adaptation methods.
Moreover, in the case of hard-label black-box domain adaptation, where only predicted labels are available, \ours~achieves significant improvements over these methods.
Code will be available at \url{https://github.com/tim-learn/ProDDing/}.

\end{abstract}

}

\maketitle
\IEEEdisplaynontitleabstractindextext
\IEEEpeerreviewmaketitle

\ifCLASSOPTIONcompsoc
\IEEEraisesectionheading{\section{Introduction}\label{sec:introduction}}
\else
\section{Introduction}
\label{sec:introduction}
\fi

Deep neural networks have achieved remarkable success across various tasks with the help of massive labeled datasets. 
However, collecting sufficient labeled data for each new task is often expensive and inefficient.
To address this challenge, transfer learning has garnered significant attention \cite{pan2009survey,zhuang2020comprehensive}, particularly in the realm of unsupervised domain adaptation (UDA) \cite{ben2007analysis,csurka2017domain}.
UDA leverages one or more related but distinct labeled datasets (source domains) to assist in recognizing unlabeled instances in a new dataset, referred to as the target domain.
In recent years, UDA methods have been extensively applied to various computer vision tasks, including image classification \cite{ganin2016domain,tzeng2017adversarial,long2018conditional}, semantic segmentation \cite{zhang2017curriculum,tsai2018learning,hu2021adversarial}, and object detection \cite{chen2018domain,khodabandeh2019robust,saito2019strong}.

\begin{figure}[!ht]
\centering
\includegraphics[width=0.88\linewidth, trim=20 10 20 10,clip]{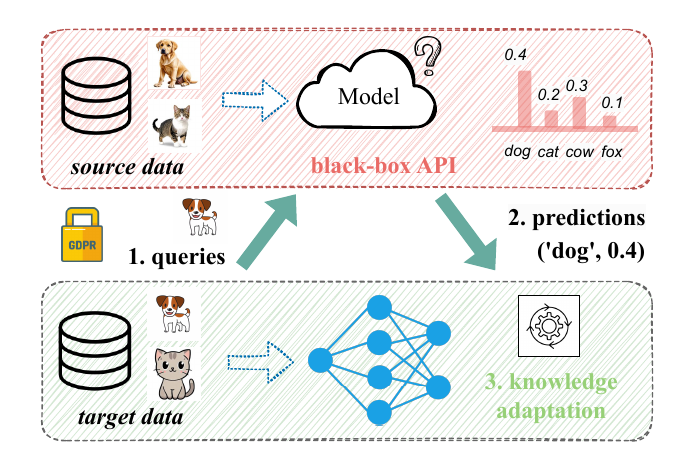}
\caption{For black-box domain adaptation, the source vendor provides only black-box predictors (e.g., through a cloud API service) to the target user, who possesses certain unlabeled data. During adaptation, only the predicted labels and their associated confidence values are accessible for target queries. When confidence values are unavailable, we refer to this scenario as \textbf{hard-label} black-box domain adaptation.}
\label{fig:framework}
\vspace{-10pt}
\end{figure}

Existing UDA methods typically require access to raw source data and rely on techniques such as domain adversarial training \cite{tzeng2017adversarial,ganin2016domain} or maximum mean discrepancy minimization \cite{tzeng2014deep,long2015learning} to align source and target features. 
However, in many cases, such as handling personal medical records, raw source data can not be shared due to privacy concerns.
To address this limitation, recent studies \cite{liang2020we,li2020model,kundu2020universal,liang2021source} have explored source-free unsupervised domain adaptation (SFUDA) by using trained source models as supervision instead of raw data, achieving promising adaptation results. 
Nevertheless, these SFUDA methods often require the source models to be carefully trained and fully disclosed to the target domain, raising two critical concerns.
First, model inversion attacks \cite{fredrikson2015model,kurmi2021domain} can potentially reconstruct raw source data, risking individual privacy.
Second, these approaches typically train source models with specialized techniques while assuming an identical target model architecture, which is especially impractical for resource-constrained users.
Thus, this paper focuses on a realistic and challenging scenario for UDA, where the source model is trained without bells and whistles and provided to the unlabeled target domain as a black-box predictor.

To illustrate this process more clearly, as shown in Fig.~\ref{fig:framework}, the target user accesses the API service provided by the source vendor to obtain the predicted label and its confidence value (\ie, the highest soft-max probability) for each instance (\eg, (`dog', 0.4)), using this information for knowledge adaptation in the unlabeled target domain.
This UDA scenario offers greater flexibility for cross-domain knowledge transfer, as it does not necessitate any specialized design for the source model.
To further mitigate privacy risks, this paper also considers the hard-label scenario, where the associated confidence value is unavailable for each query.
To tackle this challenging black-box scenario, knowledge distillation \cite{hinton2015distilling,zhao2022decoupled} offers a potential solution, wherein the target model (student) is traditionally trained by mimicking the comprehensive outputs of the source model (teacher) using labeled data. 
However, this becomes impractical in black-box UDA due to the simultaneous absence of labeled data and the inability to obtain complete teacher outputs.

In this paper, we propose a novel knowledge adaptation framework named Prototypical Distillation and Debiased Tuning (\ours).
\ours~follows a simple two-step pipeline: first, it primarily distills knowledge from the predictions of the source model, and then fine-tunes the distilled model using the unlabeled target data.
To fully exploit the confidence value provided by the source vendor, we elegantly devise an adaptive label smoothing technique that combines one-hot training labels with uniform label vectors, weighted adaptively.
Given the inherent noise in source-predicted labels, we further utilize representative prototypes---the feature centroids of classes in the target domain---and use the feature distance to these prototypes as a complementary source of supervision.
In addition to these two point-wise supervisions, we introduce two new structural regularizations into distillation: interpolation consistency training \cite{verma2019interpolation}---which ensures that predictions for interpolated samples align with their interpolated labels, and mutual information maximization \cite{liang2020we,hu2017learning}---which helps increase the diversity among target predictions.

To extract knowledge from unlabeled data, we fine-tune the distilled model in the second step, drawing inspiration from the semi-supervised method \cite{sohn2020fixmatch}.
To alleviate class-sampling bias during the weak-to-strong consistency objective, we propose to adjust the logits for each class based on estimated label frequencies.
Specifically, we introduce large offset values to the logits of dominant classes to reduce their influence, while reapplying mutual information maximization to the weakly augmented samples at the same time.
Extensive results on standard benchmark datasets (\eg, Office, Office-Home, and DomainNet) verify that \ours~consistently outperforms previous black-box UDA methods.
Furthermore, even in the challenging hard-label adaptation scenario, \ours~yields surprisingly promising performance.

Our contributions can be summarized as follows:
\begin{itemize}
\item We introduce a realistic and challenging UDA setting, called black-box domain adaptation, where the source model is limited to providing only the predicted label and, optionally, its confidence score for each target query.
\item We propose \ours, a simple yet effective framework that first distills noisy knowledge by querying the source model and then performs self-tuning solely on unlabeled data.
\item We design an adaptive label-smoothing strategy for source predictions and develop a novel unsupervised distillation method by integrating structural regularizations into the distillation process
\item We present a novel unsupervised fine-tuning strategy that mitigates class bias through logit adjustment and mutual information maximization.
\item Empirical results across diverse benchmarks confirm the superiority of \ours~over existing black-box UDA methods. Notably, when only limited information (\ie, hard labels) is available from the source vendor, \ours~again achieves the best performance.
\end{itemize}

This paper extends our previous conference publication \cite{liang2022dine} mainly in five aspects:
(1). In the distillation step, we incorporate a novel teacher supervision signal based on prototypes derived from the target data.
(2). In the self-tuning step, we implement a logit-adjustable weak-to-strong augmentation strategy to reduce class bias.
(3). In the introduced black-box setting, we are the first to investigate the hard-label case, where only the predicted labels are available from the source vendor.
(4). We also broaden the experimental evaluation by including additional datasets, such as DomainNet (large-scale) and Office-Home-RSUT (reverse label shift). 
(5). We offer a more detailed analysis to evaluate the proposed approach, with a particular focus on the newly added components in the framework.


\section{Related Work}
\subsection{Unsupervised Domain Adaptation}
Domain adaptation, a common scenario in transfer learning \cite{pan2009survey}, involves using labeled data from one or more source domains to address tasks in a related target domain with covariate shifts. 
Specifically, much of the research efforts have been devoted to unsupervised domain adaptation (UDA), where no labeled data is available in the target domain.
At early times, researchers address this problem via instance weighting \cite{huang2006correcting,sugiyama2007direct}, feature transformation \cite{pan2010domain,liang2018aggregating}, and feature space \cite{gong2012geodesic,fernando2013unsupervised,sun2016return}.
Over the past decade, deep domain adaptation methods, driven by advances in representation learning, have become prevalent and achieved remarkable progress.
To bridge the gap between features across different domains, deep UDA methods commonly employ domain adversarial learning \cite{ganin2016domain,tzeng2017adversarial,long2018conditional,hoffman2018cycada} and discrepancy minimization \cite{tzeng2014deep,long2017deep,koniusz2017domain,kang2019contrastive}.
Another branch of deep UDA methods \cite{saito2018maximum,chen2019domain,cui2020towards,jin2020minimum} focuses on the network outputs, introducing various regularization terms (\eg, entropy minimization) to achieve implicit domain alignment.
Moreover, researchers explore other aspects of neural networks for deep UDA, such as domain-specific normalization methods \cite{maria2017autodial,chang2019domain} and feature regularization techniques \cite{xu2019larger,chen2019transferability}.

Several UDA methods \cite{xie2018learning,pan2019transferrable} address class-level domain discrepancies by aligning class prototypes across domains. 
Alternatively, some approaches learn the domain-invariant features by promoting the alignment of target data with source class prototypes \cite{tanwisuth2021prototype}. 
To ensure semantic consistency in adversarial alignment, class prototypes are incorporated as conditional signals into features before being fed to the discriminator \cite{hu2021adversarial}. 
Moreover, prototypes can serve as centroids in nearest-centroid classifiers to assign pseudo-labels to unlabeled target data \cite{zhang2019category,liang2021domain}.
A closely related work to ours is \cite{zhang2021prototypical}, which utilizes distances to prototypes to iteratively refine pseudo-labels via a multiplication mechanism. 
In contrast, we introduce a straightforward weighted average of source predictions and derived prototypical pseudo-labels as the initial teacher signal.

\subsection{Source-free Domain Adaptation}
Motivated by hypothesis transfer learning \cite{kuzborskij2013stability,wang2016learning} and early parameter adaptation techniques \cite{joachims1999transductive,duan2009domain}, several pioneering studies \cite{chidlovskii2016domain,liang2019distant} have proposed shallow domain adaptation methods in the absence of source data (features).
More recently, several works \cite{liang2020we,liang2021source,kundu2020universal,li2020model} introduce the source data-free setting for deep UDA, where the source domain merely offers a trained model rather than the raw source data.
Specifically, \cite{liang2020we,liang2021source} freeze the classifier module in the provided model and fine-tune the feature module via information maximization and pseudo-labeling in the target domain. 
By contrast, \cite{li2020model} leverages a conditional generative adversarial net and incorporates generated images into the adaptation process.
This source-free paradigm \cite{li2024comprehensive} is more privacy-preserving and flexible compared to the conventional UDA setting, rapidly gaining attention among researchers in various transfer learning applications, such as semantic segmentation \cite{liu2021source,kundu2021generalize}, object detection \cite{li2021free,li2022source}, and medical image analysis \cite{bateson2022source,yang2022source}.

According to the taxonomy outlined in a recent survey \cite{liang2024comprehensive}, existing source-free domain adaptation methods can be broadly categorized into four groups: pseudo-labeling \cite{liang2020we,wang2022exploring,ding2023proxymix,litrico2023guiding}, consistency training \cite{fleuret2021uncertainty,zhang2022divide,chen2022contrastive,lee2023feature}, clustering-based training \cite{qiu2021source,xia2021adaptive,xia2022privacy,roy2022uncertainty}, and source distribution estimation \cite{li2020model,kurmi2021domain,nayak2021mining,hou2021visualizing,zhang2022divide}.
However, exposing the parameters of the trained source model in source-free domain adaptation can pose a risk due to model inversion attacks \cite{fredrikson2015model, zhang2020secret}, particularly in methods that rely on source distribution estimation.
Typically, the target network architecture is assumed to be the same as the pre-trained source one, limiting the flexibility in resource-constrained scenarios.
To thoroughly evaluate effectiveness under label shift, we also conduct experiments in partial-set and imbalanced cases, as used in previous source-free methods \cite{liang2020we,li2021imbalanced}.

\subsection{Black-box Domain Adaptation}
To enhance privacy-preserving capabilities in source-free domain adaptation, several recent works \cite{liang2022dine, zhang2021unsupervised, liu2022self} propose using the source model as a black-box predictor, where only predictions for target queries are accessible, without any knowledge of the source model's parameters.
In contrast to methods \cite{zhang2021unsupervised,liu2022self} that leverage the complete probability distribution from the source model, our previous work \cite{liang2022dine} proposes using truncated probabilities, such as the largest probability value and its associated label.
Such incomplete source predictions are more realistic in real-world APIs and have been widely adopted in subsequent black-box UDA studies \cite{chen2023classifier,yang2023divide,peng2023rain,xia2024separation,zhang2024reviewing}.
In addition to the predicted label and its probability as considered in \cite{liang2022dine}, this extension further explores a more challenging scenario where only the predicted label is available. 
Note that, we focus exclusively on image classification and do not address black-box adaptation methods for other tasks \cite{xu2022leveraging,wang2023black,cuttano2023cross,wang2024curriculum,luo2024crots,ren2024single}.

Faced with a black-box source model, a pioneering work \cite{liang2021source} partitions the target dataset into two splits and employs semi-supervised learning to enhance the performance of the uncertain split.
This divide-and-learn strategy is further employed in other methods \cite{liu2022self,yang2023divide,chen2023classifier,xia2024separation}, where the domain gap between the two splits is addressed using adversarial training \cite{chen2023classifier}, discrepancy minimization \cite{liu2022self}, or graph alignment \cite{xia2024separation}.
Additionally, \cite{zhang2021unsupervised} introduces an iterative noisy label learning approach to refine source predictions, while \cite{zhang2023black} develops a sophisticated memory mechanism to capture representative information during adaptation. 
Furthermore, \cite{peng2023rain} emphasizes consistency under both data and model variations.
\cite{shi2023source} takes a different approach by leveraging third-party data and adversarial training to train the target model.
\cite{lipton2018detecting} proposes a black-box solution for prior shifts that relies on a hold-out source set to estimate the class confusion matrix, which is sometimes hard to satisfy in practice.
Several recent studies \cite{xiao2024adversarial,tian2024clip} even leverage additional vision-language models \cite{radford2021learning} to enhance the performance of black-box domain adaptation.

\subsection{Knowledge Distillation}
Knowledge distillation \cite{gou2021knowledge} is a well-studied technique aimed at transferring knowledge from one model (commonly referred to as the teacher) to another model (the student), typically from a larger model to a smaller one.
A seminal work \cite{hinton2015distilling} shows that, augmenting the training of the student with a distillation loss, matching the predictions between teacher and student, is beneficial.
Recently, \cite{kim2021self} introduces self-knowledge distillation, showing that past predictions within the same neural network can serve as the teacher. 
Beyond supervised training, self-distillation can be effectively applied with unlabeled data in semi-supervised learning. 
For example, \cite{laine2016temporal} proposes ensembling predictions during training by using outputs from a single network across different epochs as a teacher for the current epoch. 
In contrast to maintaining an exponential moving average (EMA) prediction \cite{laine2016temporal}, \cite{tarvainen2017mean} utilizes an average of consecutive student models (past model weights) as a stronger teacher, though this approach is not suitable for black-box UDA.
In this paper, we propose an adaptive label smoothing technique on source predictions and for the first time introduce structural regularizations \cite{verma2019interpolation,gomes2010discriminative} into unsupervised distillation.

\begin{figure*}[t]
\centering
\includegraphics[width=0.85\linewidth, trim=30 10 25 10,clip]{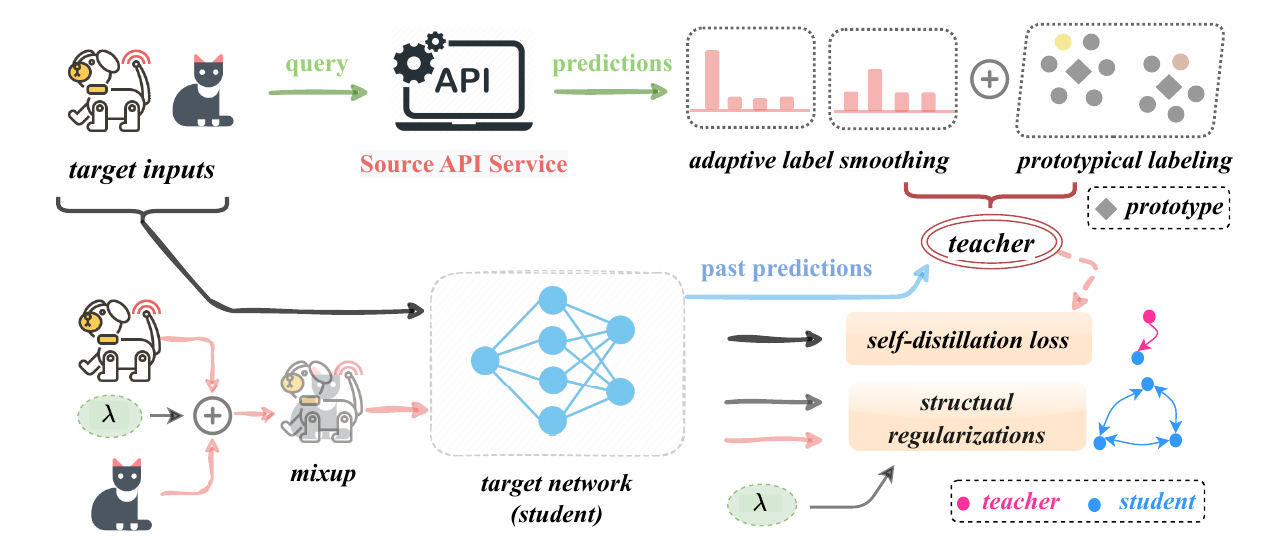}
\caption{An overview of the proposed \textbf{ProD}, the distillation step of \ours, is illustrated. 
The black-box source predictor (e.g., an API service) is used solely to initialize the memory prediction bank, which stores predictions for each target instance. 
Building on these predictions, we further employ adaptive label smoothing and prototypical pseudo-labeling to update the memory prediction bank. 
In the self-distillation process, the memory bank acts as a teacher by maintaining an exponential moving average (EMA) of predictions. 
Additionally, structural regularizations, capturing batch-wise and pair-wise data structures, are incorporated to enhance adaptation.}
\label{fig:prod}
\end{figure*}

\subsection{Semi-supervised Learning}
A senior semi-supervised learning approach \cite{lee2013pseudo} assigns pseudo-labels based on model predictions for unlabeled data, which are then used alongside labeled data to retrain the model.
Another classic method \cite{grandvalet2004semi} minimizes the entropy of each unlabeled data point as a form of regularization.
In the deep learning era, \cite{berthelot2019mixmatch} unifies existing dominant approaches for semi-supervised learning into a holistic method, which has gained increasing popularity due to its superior performance.
The holistic method is further enhanced in \cite{berthelot2020remixmatch} by incorporating distribution alignment, which encourages the marginal distribution of predictions on unlabeled data to match that of labeled data, and augmentation anchoring, which promotes weak-to-strong consistency.
In contrast, \cite{sohn2020fixmatch} presents a simple approach that also leverages weak-to-strong consistency, but uses a pre-defined threshold to filter out high-confidence samples.
A recent notable work \cite{zhang2021flexmatch} enhances \cite{sohn2020fixmatch} by introducing a curriculum learning approach that flexibly adjusts thresholds for different classes.
In this paper, we incorporate the marginal distribution of unlabeled data in weak-to-strong consistency by using logit adjustment and information maximization to mitigate class bias.

\section{Methodology}
In this paper, we focus on the $K$-way cross-domain image classification task and address a realistic yet challenging UDA setting, where only the predictions of a black-box source model are accessible for unlabeled target domain data.
For the single-source UDA scenario, the source domain $\{x_s^i, y_s^i\}_{i=1}^{n_s}$ consists of $n_s$ labeled instances, where $x_s^i \in \mathcal{X}_s, y_s^i \in \mathcal{Y}_s$, and the target domain $\{x_t^i, y_t^i\}_{i=1}^{n_t}$ consists of $n_t$ unlabeled instances, where $x_t^i \in \mathcal{X}_t, y_t^i \in \mathcal{Y}_t$, and the goal of UDA is typically to infer the values of $\{y_t^i\}_{i=1}^{n_t}$.
The label spaces are assumed to be identical across domains, \ie, $\mathcal{Y}_s = \mathcal{Y}_t$, even when label shift occurs \cite{tan2020class,li2021imbalanced}.
By contrast, partial-set UDA \cite{cao2018partial,liang2020balanced} assumes that some source classes do not exist in the target domain, \ie, $\mathcal{Y}_s \supset \mathcal{Y}_t$.
Concerning the black-box adaptation setting, only the trained source model is provided through an API service, without requiring access to the source data. 
It differs from prior source-free domain adaptation methods \cite{liang2020we,li2020model} in requiring no details about the source model, \eg, backbone type and network parameters.
In particular, \emph{only the predicted labels along with their associated probability values of the target instances $\mathcal{X}_t$ from the source model $f_s: \mathcal{X}_s\to \mathcal{Y}_s$} are utilized for knowledge adaptation in the target domain.

\subsection{Source Model Generation}
We elaborate on how to obtain the trained model from the source domain as follows. 
Unlike most source-free domain adaptation methods \cite{liang2020we,liang2021source} that elegantly design the source model with a bottleneck layer and weight normalization \cite{salimans2016weight}, we simply insert a single linear fully-connected (FC) layer after the backbone feature network and use label smoothing (LS) \cite{szegedy2016rethinking} to train $f_s$,
\begin{equation}
 \mathcal{L}_{s}(f_s;\mathcal{X}_s,\mathcal{Y}_s) = 
 \mathbb{E}_{(x_s,y_s)\in \mathcal{X}_s \times \mathcal{Y}_s} \mathcal{H}(q_s, \delta(f_s(x_s))),
 \label{eq:ls}
\end{equation} 
where $q_s=(1-\epsilon)\mathbf{1}_{y_s} + \epsilon/K$ is the smoothed label vector and $\epsilon$ is empirically set to 0.1, and $\mathbf{1}_{j}$ denotes a $K$-dimensional one-hot encoding with only the $j$-th value being 1.
Moreover, $\delta(\cdot)$ denotes the softmax function, and $\mathcal{H}(q, p)$ denotes the cross-entropy between $p$ and $q$.

\textbf{Remark \#1.} In contrast, for the self-defined target network $f_t: \mathcal{X}_t\to \mathcal{Y}_t$, we adopt the common practice in source-free domain adaptation \cite{liang2020we,liang2021source,qu2022bmd,yang2023trust,mitsuzumi2024understanding}.
Specifically, the bottleneck layer consists of a batch normalization layer and an FC layer, while the classifier includes a weight normalization layer followed by an FC layer.

\subsection{Prototypical and Adaptive Knowledge Distillation}
To extract knowledge from a black-box model, a natural solution is knowledge distillation \cite{hinton2015distilling}, which trains the target model (student) to replicate the predictions of the source model (teacher).
However, existing knowledge distillation methods are primarily designed for supervised training tasks, with the consistency loss serving as a regularization term, as shown below:
\begin{equation}
 \mathcal{L}_{kd}(f_t;\mathcal{X}_t, f_s) =
 \mathbb{E}_{x_t\in \mathcal{X}_t} D_{kl}\left(\delta (f_s(x_t)) \ ||\ \delta(f_t(x_t))\right),
 \label{eq:kd}
\end{equation} 
where $D_{kl}$ denotes the Kullback-Leibler (KL) divergence loss. 
However, the source model $f_s$'s outputs for target instances are often \emph{inaccurate and sometimes even incomplete}.
For the studied black-box adaptation problem, highly relying on the teacher $f_s(x_t)$ through a consistency loss is no longer desirable.
Thus, we propose adaptively smoothing the teacher $p$ by focusing on the top-$r$ largest values as:
\begin{equation}
\text{AdaLS}(p,r)_i =\left\{
\begin{aligned}
 p_i, \qquad \qquad \qquad \qquad \qquad\; & i \in \mathcal{T}^r_p, \\
 (1 - \sum\nolimits_{j \in \mathcal{T}^r_p} p_j)/ (K-r), \; & \text{otherwise}.
\end{aligned}
\right.
\label{eq:als}
\end{equation} 	
Here $\mathcal{T}^r_p$ represents the index set of the top-$r$ classes in $p$.
We refer to the transformation in Eq.~(\ref{eq:als}) as adaptive label smoothing (\textbf{Adaptive LS}), as it retains instance-specific top-$r$ values, which vary across samples. 
Using the smoothed output ($r=1$) means that we merely need the predicted label along with its maximum probability, which sounds more flexible when using an API service provided by profitable companies.	
For simplicity, we denote the refined output as 
\begin{equation}
p_s(x_t)=\text{AdaLS}\left(\delta(f_s(x_t)),r=1\right)
\label{eq:als-2}
\end{equation}
throughout this paper.
The refined output $p_s(x_t)$ is expected to work better than the original output $p$ for several reasons: 
1) it reduces redundant and noisy information by focusing on the pseudo label (the class associated with the largest value) and applying a uniform distribution to the other classes, similar to label smoothing \cite{szegedy2016rethinking}; 
2) it does not rely solely on the noisy pseudo label, instead using the largest value as a measure of confidence, akin to self-weighted pseudo labeling \cite{iscen2019label}.

Inspired by previous studies \cite{liang2020we,zhang2021prototypical} that leverage prototypes to denoise the pseudo labels for unlabeled target data, we further develop a prototypical pseudo-labeling strategy based on the source predictions. 
Once we have a pre-trained feature extractor $g_t$, the prototype of the $k$-th class in the target domain could be calculated as follows, 
\begin{equation}
\mathcal{C}_k = \frac{\sum_{x_t\in \mathcal{X}_t} [p_s(x)]_k\, g_t(x_t)}{\sum_{x\in \mathcal{X}_t}[p_s(x)]_k}.
\label{eq:proto}
\end{equation}
We can then obtain the soft pseudo labels as the softmax over the distance between the target features and the prototypes as follows:
\begin{equation}
[p_s^t(x_t)]_k = \frac{\exp\left(-d(g_t(x_t), \, \mathcal{C}_k)/\tau\right)}{\sum_{k'} \exp\left((-d(g_t(x_t), \, \mathcal{C}_{k'})/\tau\right)},
\label{eq:proto-pl}
\end{equation}
where $\tau$ represents the temperature parameter, which is empirically set to 0.1, and $d(\cdot,\cdot)$ denotes the cosine distance.
To integrate these two types of pseudo labels in Eq.~(\ref{eq:als-2}) and Eq.~(\ref{eq:proto-pl}), we present a simple weighted addition below,
\begin{equation}
P_s^t(x_t) = \beta\, p_s(x_t) + (1-\beta)\, p_s^t(x_t) ,
\label{eq:init}
\end{equation}
where the balancing parameter $\beta \in [0,1]$ is empirically set to 0.5. The prototypical pseudo label is omitted when $\beta=1$. 

To further alleviate the noise in the teacher prediction, we follow \cite{laine2016temporal,kim2021self} and adopt a self-distillation strategy, shown in Fig.~\ref{fig:prod}, maintaining an EMA prediction by 
\begin{equation}
 P_s^t(x_t) \gets \gamma\, P_s^t(x_t) + (1-\gamma)\, \delta(f_t(x_t)), \ \forall x_t \in \mathcal{X}_t,
 \label{eq:ema}
\end{equation}
where $\gamma$ is a momentum hyper-parameter, which is empirically set to 0.7.
Following \cite{laine2016temporal}, we update teacher predictions after every training epoch.
When $\gamma=1$, there exists no temporal ensembling, \ie, the refined source predictions keep acting as a teacher throughout distillation.

\textbf{Remark \#2.} 
To calculate the prototypical soft pseudo labels, features are reduced through principal component analysis and $l_2$-normalized.
To determine the feature extractor $g_t$, we could use powerful large neural networks; however, for a fair comparison with other methods, we default to utilizing the feature encoder module in the distilled target network $f_t$. 
The analysis is provided in the experiments.

\subsection{Self-distillation with Structural Regularizations}
As mentioned earlier, the teacher output from the source model is likely to be inaccurate and noisy due to domain shift. 
Even though we propose a promising solution in Eq.~(\ref{eq:init}), which only considers point-wise information during the distillation process, it fails to account for the data structure in the target domain, making it insufficient for effective noisy knowledge distillation. 
To address this, we incorporate the structural information in the target domain to regularize the distillation process.
On the one hand, we consider the pairwise structural information via MixUp \cite{zhang2017mixup}, and employ the interpolation consistency training \cite{verma2019interpolation} technique as below,
\begin{equation}
\begin{aligned}
\mathcal{L}_{mix} & (f_t;\mathcal{X}_t) =
\mathbb{E}_{x^t_i, x^t_j\in \mathcal{X}_t} \mathbb{E}_{\lambda \in \text{Beta}(\alpha,\alpha)} \\
& \mathcal{H}\left(
\text{Mix}_{\lambda}\left(\delta(f'_t(x^t_i)), \delta(f'_t(x^t_j))\right), \delta(f_t\left(\text{Mix}_{\lambda}(x^t_i, x^t_j)\right))
\right), \\
\end{aligned}
\label{eq:mix}
\end{equation}
where the operation $\text{Mix}_{\lambda}(a,b)=\lambda\cdot a + (1-\lambda)\cdot b$ denotes the MixUp operation, $\lambda$ is sampled from a Beta distribution, and $\alpha$ is the hyper-parameter, empirically set to 0.3 according to \cite{zhang2017mixup}.
Note that $f'_t$ just offers the values of $f_t$ but requires no gradient optimization. 
Here we do not adopt the EMA update strategy in \cite{verma2019interpolation} for $f'_t$.
Eq.~(\ref{eq:mix}) can be treated to augment the target domain with more interpolated samples, which is beneficial for better generalization ability. 

On the other hand, we also consider the global structural information during distillation in the target domain.
In fact, during distillation, the classes with a large number of instances are relatively easy to learn, which may wrongly recognize some confusing target instances as such classes in turn. 
To circumvent this problem, we attempt to encourage diversity among the predictions of all the target instances.
Specifically, we try to maximize the widely-used mutual information objective \cite{gomes2010discriminative,hu2017learning,liang2020we,xie2021muscle} in the following,
\begin{equation}
 \begin{aligned}
 \mathcal{L}_{mi}(f_t;\mathcal{X}_t) &= 
 \mathcal{H}(\mathcal{Y}_t) - \mathcal{H}(\mathcal{Y}_t | \mathcal{X}_t) \\
 & = h\left(\mathbb{E}_{x_t\in\mathcal{X}_t} \delta(f_t(x_t))\right) - \mathbb{E}_{x_t\in\mathcal{X}_t}\ h\left(\delta(f_t(x_t))\right), 
 \end{aligned}
 \label{eq:mi}
\end{equation}
where $h(p)=-\sum_i p_i \log p_i$ represents the conditional entropy function.
Note that, increasing the marginal entropy $\mathcal{H}(\mathcal{Y}_t)$ encourages the label distribution to be uniform while decreasing the conditional entropy $\mathcal{H}(\mathcal{Y}_t|\mathcal{X}_t)$ encourages unambiguous network predictions.

By combining the objectives defined in Eqs.~(\ref{eq:kd}), (\ref{eq:mix}), and (\ref{eq:mi}), the final loss function for the first distillation step of \ours~is formulated as follows:
\begin{equation}
\mathcal{L}_{prod} = \mathcal{L}_{skd} + \mathcal{L}_{mix} - \mathcal{L}_{mi},\\
\label{eq:prod}
\end{equation}
where $\mathcal{L}_{skd}(f_t;\mathcal{X}_t, f_s) = \mathbb{E}_{x_t\in \mathcal{X}_t} D_{kl}\left(\delta(P_s^t(x_t)) \ ||\ \delta(f_t(x_t))\right)$, and both structural regularizations equally contribute to the \textbf{Pro}totypical \textbf{D}istillation method (\textbf{ProD}).
Unlike the closely related work \cite{zhang2021unsupervised}, which iteratively refines pseudo labels and optimizes the target network, ProD adopts a unified approach to directly learn accurate predictions for the target data, which is more effective at capturing the inherent data structure of the target domain.

\begin{figure}[t]
\centering
\includegraphics[width=0.9\linewidth, trim=20 10 25 10,clip]{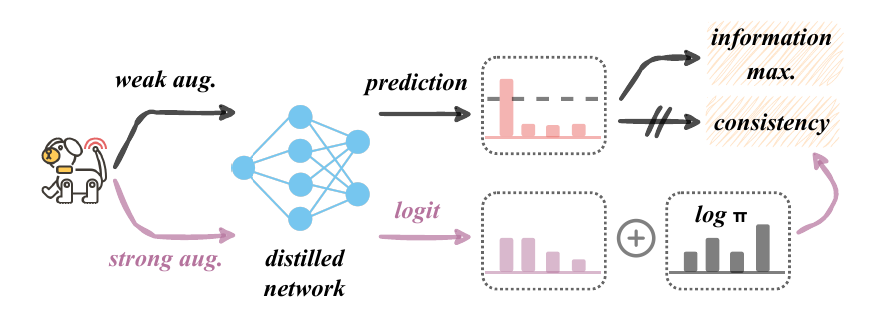}
\caption{An overview of the proposed \textbf{Ding}, the fine-tuning step of \ours, is illustrated. 
Built on the network distilled in the first step, we pursue weak-to-strong consistency with a pre-defined threshold over the prediction of the weak augmented sample. 
In addition to mutual information maximization, we adjust the logits of the strong version to mitigate class bias, where $\pi$ denotes the estimate of the class priors.}
\label{fig:ding}
\end{figure}

\subsection{Debiased Fine-tuning}
Through the proposed prototypical structural knowledge distillation method from black-box source predictors $f_s$, it is expected to learn a well-performing white-box target model. 
However, the distilled model seems sub-optimal since it is mainly optimized via the point-wise knowledge distillation term in Eq.~(\ref{eq:kd}), which highly depends on the source predictions.
Inspired by DIRT-T \cite{shu2018dirt}, we hypothesize that the network performance can be further improved by introducing a secondary training phase focused exclusively on minimizing violations of the target-side cluster assumption. 
Instead of using the parameter-sensitive virtual adversarial training \cite{shu2018dirt}, we refine the distilled target model by adopting the widely-used weak-to-strong consistency technique introduced in FixMatch \cite{sohn2020fixmatch}, as illustrated below:
\begin{equation}
\mathcal{L}_{fm} = \mathbb{E}_{x_t\in\mathcal{X}_t} \ \mathbb{I}(max(\delta(f_t(x_t)))\geq \eta)\ \mathcal{H}(\mathbf{1}_{\hat{y}_t}, \delta (f_t(\mathcal{A}(x_t)))),
\label{eq:fixmatch}
\end{equation}
where $\hat{y}_t=\arg\max(\delta(f_t(x_t)))$ represents the hard pseudo-label based on weak augmentation, $\eta$ is the threshold (dashed in Fig.~\ref{fig:ding}), and $\mathcal{A}(\cdot)$ denotes the augmentation sampled from AutoAugment \cite{cubuk2019autoaugment}.

As mentioned earlier, class bias is a common obstacle in the unsupervised learning process. 
Inspired by \cite{menon2021long}, which adjusts the logits per class for long-tail learning, we incorporate an estimate of the class prior into the consistency loss to mitigate the bias toward `easy' classes.
Firstly, the class prior is iteratively estimated per epoch by
\begin{equation}
\pi_k = \frac{\mathbb{E}_{x_t\in\mathcal{X}_t}\,\mathbb{I}(\hat{y}_t = k)}{n_t}, \, k \in [1, \dots, K].
\label{eq:label-prior}
\end{equation}
Secondly, we adjust the logits of samples under strong augmentations, as shown in Fig.~\ref{fig:ding}. 
The adjustment is formulated as follows:
\begin{equation}
\begin{aligned}
\mathcal{L}_{afm} = \mathbb{E}_{x_t\in\mathcal{X}_t}& \ \mathbb{I}(max(\delta(f_t(x_t)))\geq \eta)\ \\
&\mathcal{H}(\mathbf{1}_{\hat{y}_t}, \delta(f_t(\mathcal{A}(x_t)) + \rho \log \pi)),
\end{aligned}
\label{eq:logit}
\end{equation}
where $\rho$ denotes the adjustable parameter, empirically set to 0.5. 
At the same time, we can also apply mutual information maximization, as described in Eq.~(\ref{eq:mi}), to alleviate class bias in samples under weak augmentation.
Finally, the overall loss for the second step of \ours~(referred to as \textbf{D}ebiased Fine-tun\textbf{ing}, \textbf{Ding}) is given by:
\begin{equation}
\mathcal{L}_{ding} = \mathcal{L}_{afm} - \mathcal{L}_{mi}.\\
\label{eq:ding}
\end{equation}

\begin{algorithm}[!tb]
\caption{Pseudocode of \ours~for black-box UDA.}
\small 
\setstretch{0.99}
\label{alg:prodding}
\begin{algorithmic}
 \STATE {\bfseries \textcolor{Gray}{1. Source Model Generation}}
 \STATE {\bfseries Require:} $\{x_s^i,y_s^i\}_{i=1}^{n_s}$.
 \STATE $\triangleright$ Train $f_s$ via minimizing the objective in Eq.~(\ref{eq:ls}).
 \STATE {\bfseries \textcolor{Gray}{2. Prototypical Distillation}} 
 \STATE {\bfseries Require:} Target data $\{x_t^i\}_{i=1}^{n_t}$, source predictions $\{p_s(x_t)\}_{i=1}^{n_t}$, parameters $\beta=0.5,\tau=0.1$, the number of epochs $T_m$.
 \STATE $\triangleright$ Obtain the smoothed source predictions via Eq.~(\ref{eq:als-2}) \textbf{($r=1$)}.
 \STATE $\triangleright$ Obtain the prototypical pseudo labels via Eq.~(\ref{eq:proto-pl}).
 \STATE $\triangleright$ Initialize the teacher output $P_s^t(x_t)$ via Eq.~(\ref{eq:init}).
 \FOR {$e=1$ {\bfseries to} $T_m$}
 \FOR{$i=1$ {\bfseries to} $n_b$}
 \STATE $\triangleright$ Sample a batch from target data.
 \STATE $\triangleright$ Apply MixUp within the batch.
 \STATE $\triangleright$ Update $f_t$ via minimizing the objective in Eq.~(\ref{eq:prod}).
 \ENDFOR
 \STATE $\triangleright$ Update the teacher output $P_s^t(x_t)$ via Eq.~(\ref{eq:ema}).
 \ENDFOR
 \STATE {\bfseries \textcolor{Gray}{3. Debiased Fine-tuning}}
 \STATE {\bfseries Require:} Target data $\{x_t^i\}_{i=1}^{n_t}$, parameters $\rho=0.5,\eta$, the distilled target network $f_t$, the number of epochs $T_m$.
 \FOR {$e=1$ {\bfseries to} $T_m$}
 \STATE $\triangleright$ Obtain the pseudo labels $\hat{y}_t$ under weak augmentation.
 \STATE $\triangleright$ Update the label prior estimate $\pi$ using Eq.~(\ref{eq:label-prior}).
 \FOR{$i=1$ {\bfseries to} $n_b$}
 \STATE $\triangleright$ Sample a batch from target data with both weak and strong augmentations.
 \STATE $\triangleright$ Update $f_t$ via minimizing the objective in Eq.~(\ref{eq:ding}).
 \ENDFOR
 \ENDFOR
\end{algorithmic}
\end{algorithm}

So far, we have presented all the details of the two steps within the proposed framework (\ours). A full description of \ours~can be found in Algorithm~\ref{alg:prodding}.

\textbf{Remark \#3.} In a more challenging case, i.e., hard-label black-box UDA, where only the predicted label is available for each target query, we simply employ the conventional label smoothing technique instead of AdaLS in Eq.~(\ref{eq:als-2}).

\begin{table*}[!ht]
\caption{Accuracies (\%) on the \textbf{Office-Home} dataset for UDA under two black-box scenarios. The `Hard' setting indicates that only the predicted label is available for each target query. The best results are bolded and highlighted in different colors for each scenario.}
\label{tab:home}
\setlength{\tabcolsep}{2.5pt}
\resizebox{0.99\linewidth}{!}{
\begin{tabular}{lccccccccccccca}
\toprule
Methods & Hard & Ar$\to$Cl & Ar$\to$Pr & Ar$\to$Re & Cl$\to$Ar & Cl$\to$Pr & Cl$\to$Re & Pr$\to$Ar & Pr$\to$Cl & Pr$\to$Re & Re$\to$Ar & Re$\to$Cl & Re$\to$Pr & Avg. \\
\midrule
No Adapt. & - & 44.7 & 68.3 & 75.2 & 54.4 & 63.4 & 66.7 & 52.3 & 40.3 & 73.5 & 66.5 & 46.4 & 78.0 & 60.8 \\
NLL-OT \cite{asano2019self} & \xmark & 46.3 & 69.6 & 75.8 & 56.7 & 65.4 & 68.3 & 54.0 & 41.8 & 74.4 & 67.0 & 48.7 & 78.8 & 62.2 \\
NLL-KL \cite{zhang2021unsupervised} & \xmark & 47.1 & 70.1 & 76.1 & 57.1 & 65.9 & 68.5 & 54.2 & 42.4 & 74.5 & 67.1 & 48.9 & 79.0 & 62.6 \\
NLL-MM \cite{liang2021source} & \xmark & 47.5 & 75.9 & 80.2 & 62.0 & 74.4 & 76.4 & 57.9 & 43.6 & 80.5 & 67.7 & 47.4 & 81.8 & 66.3 \\
SHOT$^\dagger$ \cite{liang2020we} & \xmark & 53.2 & 77.5 & 80.3 & 66.3 & 77.2 & 77.7 & 62.3 & 48.6 & 81.0 & 70.8 & 54.5 & 82.5 & 69.3 \\
DINE \cite{liang2022dine} & \xmark & 54.7 & 78.9 & 81.7 & 64.4 & 75.2 & 78.4 & 62.5 & 51.0 & 81.8 & 70.9 & 57.1 & 84.9 & 70.1 \\
BETA \cite{yang2023divide} & \xmark & 56.2 & 79.7 & 82.8 & 66.5 & 76.3 & 79.1 & 64.3 & 52.2 & 82.9 & 72.4 & 58.4 & \best{85.0} & 71.3 \\
SEAL \cite{xia2024separation} & \xmark & 56.5 & 79.8 & 82.6 & \best{68.7} & 77.9 & 79.0 & 65.2 & 53.7 & 83.4 & \best{73.0} & 58.6 & 84.7 & 71.9 \\
ProD & \xmark & 53.5 & 80.0 & 81.2 & 67.8 & 78.0 & 79.3 & 63.7 & 51.1 & 82.3 & 70.8 & 55.8 & 84.1 & 70.6 \\
\ours & \xmark & \best{56.9} & \best{80.9} & \best{83.2} & 68.2 & \best{79.6} & \best{81.6} & \best{65.4} & \best{54.6} & \best{83.7} & 71.6 & \best{59.0} & 84.9 & \best{72.5} \\
\midrule
SHOT$^\dagger$ \cite{liang2020we} & \cmark & 52.0 & 76.9 & 79.2 & 64.4 & 76.1 & 75.7 & 60.4 & 47.4 & 79.3 & 69.4 & 52.8 & 81.4 & 67.9 \\
DINE \cite{liang2022dine} & \cmark & 52.4 & 75.6 & 79.3 & 62.8 & 74.6 & 75.1 & 59.5 & 48.0 & 79.0 & 69.5 & 55.4 & 82.9 & 67.8 \\
BETA \cite{yang2023divide} & \cmark & 51.6 & 75.1 & 79.4 & 62.1 & 74.3 & 75.6 & 59.1 & 48.5 & 79.1 & 69.4 & 55.1 & 82.6 & 67.7 \\
SEAL \cite{xia2024separation} & \cmark & 50.5 & 74.7 & 78.8 & 61.6 & 71.3 & 72.9 & 58.2 & 46.3 & 78.1 & 69.9 & 52.7 & 82.4 & 66.4 \\
ProD & \cmark & 51.5 & 78.9 & 80.7 & 66.6 & 77.0 & 78.5 & 62.8 & 49.1 & 81.0 & 70.7 & 54.5 & 83.3 & 69.5 \\
\ours & \cmark & \sest{55.5} & \sest{79.5} & \sest{82.6} & \sest{68.1} & \sest{79.5} & \sest{80.8} & \sest{64.3} & \sest{53.1} & \sest{82.5} & \sest{71.3} & \sest{57.6} & \sest{84.7} & \sest{71.6} \\
\bottomrule
\end{tabular}}
\end{table*}

\section{Experiments}
\subsection{Setup}
\noindent \textbf{a) Datasets.} 
\textbf{Office-Home} \cite{venkateswara2017deep} is a challenging medium-sized benchmark comprising four distinct domains: Artistic images (\textbf{Ar}), Clip Art (\textbf{Cl}), Product images (\textbf{Pr}), and Real-World images (\textbf{Re}). Each domain includes 65 categories of everyday objects.
\textbf{Office} \cite{saenko2010adapting} is a widely-used UDA benchmark for cross-domain object recognition. It includes three domains: Amazon (\textbf{A}), DSLR (\textbf{D}), and Webcam (\textbf{W}), with each domain containing 31 object classes commonly found in office environments.
\textbf{DomainNet} \cite{peng2019moment} is a large-scale dataset encompassing common objects across six diverse domains, each containing 345 categories, such as bracelets, planes, birds, and cellos. The six domains are: clipart-style illustrations (\textbf{clp}), infographic-style images (\textbf{inf}), paintings (\textbf{pnt}), simplistic drawings from the quick draw game (\textbf{qdr}), real-world photographs (\textbf{rel}), and hand-drawn sketches (\textbf{skt}).
To investigate performance under label shift for different UDA methods, we also consider two variants of \textbf{Office-Home}: \textbf{Office-Home-RSUT} \cite{tan2020class,prabhu2021sentry}, where the source and target label distributions are manually modified to be reversed versions of one another, and \textbf{Office-Home-Partial} \cite{cao2018partial,liang2020balanced}, which selects the first 25 categories (in alphabetical order) from the 65 classes in each domain as the partial target domain.

\noindent \textbf{b) Baseline methods.}
{No Adapt.} is also known as `source only' in this field that infers the class label from the source predictions. 
Throughout this paper, we compare \ours~with three existing black-box UDA methods: DINE \cite{liang2022dine}, BETA \cite{yang2023divide}, and SEAL \cite{xia2024separation}.
Besides, we extend a popular source-free UDA method, SHOT \cite{liang2020we}, to both black-box scenarios, denoted as SHOT$^\dagger$.
In particular, SHOT$^\dagger$ first learns a white-box target model by utilizing the source predictions with a weighted cross-entropy loss. Subsequently, SHOT$^\dagger$ applies the algorithm proposed in \cite{liang2020we} to adapt the learned model to the target domain.
Additionally, we construct two baselines using noisy label learning: NLL-OT and NLL-KL, which regularly update the pseudo labels during the training process with different optimization objectives.
NLL-OT adopts the optimal transport (OT) technique \cite{asano2019self}, while NLL-KL adopts the diversity-promoting KL divergence \cite{zhang2021unsupervised} to refine the noisy pseudo labels. 
In contrast, NLL-MM employs the divide-to-learn strategy \cite{liang2021source} based on confidence values and leverages the semi-supervised learning algorithm, MixMatch \cite{berthelot2020remixmatch}, to train the target model.
For our methods, we also provide the results of Prod in each table.
As the source model plays a crucial role in source-free UDA, all the results presented in the experiments were reproduced by us using the source code provided by the authors of each respective work. 
We attempted to reproduce other existing black-box UDA approaches \cite{liu2022self, peng2023rain, zhang2024reviewing}, but were unable to match the results reported in their original papers.

\begin{table}[!ht]
\caption{Accuracies (\%) on the \textbf{Office} dataset for UDA under two black-box scenarios.}
\label{tab:office}
\setlength{\tabcolsep}{2.0pt}
\resizebox{0.99\linewidth}{!}{$
\begin{tabular}{lccccccca}
\toprule
Methods & Hard & A$\to$D & A$\to$W & D$\to$A & D$\to$W & W$\to$A & W$\to$D & Avg. \\
\midrule
No Adapt. & - & 80.3 & 77.9 & 61.5 & 94.5 & 63.5 & 98.4 & 79.3 \\
NLL-OT \cite{asano2019self} & \xmark & 85.9 & 83.4 & 63.5 & 96.1 & 65.1 & 98.4 & 82.1 \\
NLL-KL \cite{zhang2021unsupervised} & \xmark & 88.4 & 83.8 & 63.9 & 96.4 & 65.4 & 98.4 & 82.7 \\
NLL-MM \cite{liang2021source} & \xmark & 84.7 & 85.5 & 69.4 & 95.8 & 72.4 & 96.3 & 84.0 \\
SHOT$^\dagger$ \cite{liang2020we} & \xmark & 93.0 & \best{92.1} & 72.8 & 95.7 & 74.0 & 97.3 & 87.5 \\
DINE \cite{liang2022dine} & \xmark & 88.8 & 89.4 & 74.8 & 97.7 & 74.8 & 98.8 & 87.4 \\
BETA \cite{yang2023divide} & \xmark & 91.3 & 88.8 & 75.4 & \best{98.3} & 76.6 & 98.7 & 88.2 \\
SEAL \cite{xia2024separation} & \xmark & 88.4 & 88.1 & 75.6 & 98.0 & 76.7 & \best{98.9} & 87.6 \\
ProD & \xmark & 94.0 & 91.2 & 74.8 & 96.0 & 75.3 & 97.2 & 88.1 \\
\ours & \xmark & \best{94.6} & 92.0 & \best{75.7} & 96.2 & \best{76.8} & 97.7 & \best{88.8} \\
\midrule
SHOT$^\dagger$ \cite{liang2020we} & \cmark & 92.3 & 91.5 & 71.7 & 95.0 & 72.8 & 97.5 & 86.8 \\
DINE \cite{liang2022dine} & \cmark & 88.5 & 87.2 & 70.0 & 97.0 & 71.3 & 98.8 & 85.4 \\
BETA \cite{yang2023divide} & \cmark & 88.4 & 87.1 & 69.8 & 96.7 & 71.2 & \sest{99.0} & 85.4 \\
SEAL \cite{xia2024separation} & \cmark & 86.8 & 85.7 & 68.5 & 96.3 & 70.0 & 98.6 & 84.3 \\
ProD & \cmark & 93.2 & 91.3 & 72.5 & 97.1 & 73.8 & 97.9 & 87.6 \\
\ours & \cmark & \sest{94.1} & \sest{92.8} & \sest{75.3} & \sest{97.7} & \sest{76.6} & 97.9 & \sest{89.1} \\
\bottomrule
\end{tabular}
$}
\end{table}

\begin{table*}[!ht]
\caption{Accuracies (\%) on the \textbf{DomainNet} dataset for UDA under two black-box scenarios. $^\circ$ denotes results under the hard-label scenario.}
\label{tab:domainnet}
\setlength{\tabcolsep}{1.8pt}
\resizebox{1.0\linewidth}{!}{
\begin{tabular}{cccccccccccccccccccccccc}
\toprule
No Adapt. & clp & inf & pnt & qdr & rel & skt & Avg. & NLL-OT \cite{asano2019self} & clp & inf & pnt & qdr & rel & skt & Avg. & NLL-KL \cite{zhang2021unsupervised} & clp & inf & pnt & qdr & rel & skt & Avg. \\
\midrule
clp$\to$ & - & 16.3 & 34.7 & 9.8 & 51.9 & 40.3 & 30.6 & clp$\to$ & -& 14.8 & 37.0 & 17.0 & 60.4 & 40.7 & 34.0 & clp$\to$ & -& 15.7 & 40.7 & 17.8 & 62.6 & 42.6 & 35.9 \\
inf$\to$ & 31.2 & - & 30.8 & 2.3 & 47.3 & 24.7 & 27.2 & inf$\to$ & 38.7 & -& 34.6 & 3.6 & 57.9 & 32.2 & 33.4 & inf$\to$ & 40.6 & -& 38.1 & 4.4& 60.0 & 33.5 & 35.3 \\
pnt$\to$ & 40.1 & 16.3 & - & 2.6 & 57.2 & 33.8 & 30.0 & pnt$\to$ & 46.3 & 15.4 & -& 5.6 & 61.4 & 38.5 & 33.4 & pnt$\to$ & 48.0 & 16.4 & -& 6.5& 63.5 & 39.8 & 34.9 \\
qdr$\to$ & 9.6 & 1.0 & 1.5 & - & 3.6 & 7.8 & 4.7 & qdr$\to$ & 17.4 & 1.2& 3.2& - & 10.0 & 13.2 & 9.0 & qdr$\to$ & 17.4 & 1.2& 3.4& -& 10.7 & 13.8 & 9.3 \\
rel$\to$ & 46.7 & 18.9 & 46.3 & 4.5 & - & 34.3 & 30.1 & rel$\to$ & 50.1 & 17.3 & 42.0 & 7.5 & -& 37.1 & 30.8 & rel$\to$ & 52.1 & 18.7 & 47.2 & 8.6& -& 38.7 & 33.1 \\
skt$\to$ & 47.9 & 12.7 & 33.6 & 11.6 & 45.9 & - & 30.4 & skt$\to$ & 50.6 & 14.2 & 39.0 & 19.6 & 59.1 & -& 36.5 & skt$\to$ & 53.8 & 15.2 & 43.1 & 20.1 & 61.2 & -& 38.7 \\
\rowcolor{blue!10}
Avg. & 35.1 & 13.0 & 29.4 & 6.2 & 41.2 & 28.2 & 25.5 & Avg. & 40.6 & 12.6 & 31.2 & 10.7 & 49.8 & 32.4 & 29.5 & Avg. & 42.4 & 13.4 & 34.5 & \best{11.5} & 51.6 & 33.7 & 31.2 \\
\midrule
NLL-MM \cite{liang2021source} & clp & inf & pnt & qdr & rel & skt & Avg. & SHOT$^\dagger$ \cite{liang2020we} & clp & inf & pnt & qdr & rel & skt & Avg. & DINE \cite{liang2022dine} & clp & inf & pnt & qdr & rel & skt & Avg. \\
clp$\to$ & - & 13.5 & 38.4 & 10.7 & 58.3 & 40.7 & 32.3 & clp$\to$ & - & 15.6 & 41.5 & 16.5 & 62.2 & 41.8 & 35.5 & clp$\to$ & - & 15.9 & 42.6 & 13.8 & 60.9 & 43.1 & 35.3 \\
inf$\to$ & 32.3 & - & 33.4 & 2.5 & 54.3 & 26.6 & 29.8 & inf$\to$ & 41.5 & - & 38.9 & 4.7 & 58.7 & 33.4 & 35.4 & inf$\to$ & 37.6 & - & 41.1 & 4.3 & 57.3 & 31.6 & 34.4 \\
pnt$\to$ & 38.9 & 14.1 & - & 2.6 & 61.0 & 32.6 & 29.8 & pnt$\to$ & 49.2 & 16.7 & - & 7.2 & 62.1 & 39.4 & 34.9 & pnt$\to$ & 44.0 & 16.3 & - & 5.5 & 62.6 & 39.3 & 33.5 \\
qdr $\to$ & 9.8 & 0.7 & 1.3 & - & 4.2 & 7.9 & 4.8 & qdr$\to$ & 20.8 & 1.9 & 3.9 & - & 11.5 & 15.2 & 10.7 & qdr$\to$ & 14.0 & 0.8 & 3.3 & - & 9.1 & 12.0 & 7.8 \\
rel$\to$ & 47.0 & 15.0 & 49.3 & 4.9 & - & 35.1 & 30.2 & rel$\to$ & 52.6 & 18.9 & 46.5 & 8.4 & - & 38.9 & 33.1 & rel$\to$ & 51.9 & 18.5 & 52.0 & 7.5 & - & 39.8 & 33.9 \\
skt$\to$ & 48.2 & 10.3 & 34.1 & 13.0 & 51.9 & - & 31.5 & skt$\to$ & 54.3 & 15.1 & 43.6 & 18.1 & 60.3 & - & 38.3 & skt$\to$ & 52.8 & 14.5 & 44.0 & 16.7 & 58.2 & - & 37.2 \\
\rowcolor{blue!10}
Avg. & 35.2 & 10.7 & 31.3 & 6.7 & 45.9 & 28.6 & 26.4 & Avg. & 43.7 & \best{13.6} & 34.9 & 11.0 & 51.0 & 33.8 & 31.3 & Avg. & 40.1 & 13.2 & 36.6 & 9.5& 49.6 & 33.2 & 30.4 \\
\midrule
BETA \cite{yang2023divide} & clp & inf & pnt & qdr & rel & skt & Avg. & SEAL \cite{xia2024separation} & clp & inf & pnt & qdr & rel & skt & Avg. & ProD & clp & inf & pnt & qdr & rel & skt & Avg. \\
clp$\to$ & - & 12.8 & 39.9 & 12.2 & 61.0 & 38.9 & 33.0 & clp$\to$ & - & 15.4 & 42.6 & 12.0 & 63.6 & 43.9 & 35.5 & clp$\to$ & - & 15.7 & 44.2 & 13.3 & 63.7 & 41.9 & 35.8 \\
inf$\to$ & 34.0 & - & 39.2 & 3.4 & 57.6 & 27.8 & 32.4 & inf$\to$ & 39.8 & - & 42.6 & 3.5 & 59.0 & 32.1 & 35.4 & inf$\to$ & 38.4 & - & 43.7 & 4.0 & 60.7 & 31.7 & 35.7 \\
pnt$\to$ & 38.5 & 14.2 & - & 3.0 & 62.6 & 36.5 & 30.9 & pnt$\to$ & 45.1 & 16.6 & - & 3.8 & 64.0 & 41.0 & 34.1 & pnt$\to$ & 44.2 & 15.9 & - & 5.0 & 64.3 & 38.3 & 33.5 \\
qdr$\to$ & 10.4 & 0.7 & 1.5 & - & 7.5 & 8.7 & 5.7 & qdr$\to$ & 16.1 & 1.2 & 3.5 & - & 9.1 & 12.8 & 8.5 & qdr$\to$ & 18.6 & 0.9 & 7.1 & - & 16.8 & 13.4 & 11.3 \\
rel$\to$ & 46.6 & 14.7 & 49.5 & 6.1 & - & 36.3 & 30.6 & rel$\to$ & 54.1 & 19.0 & 52.7 & 6.0 & - & 41.8 & 34.7 & rel$\to$ & 51.8 & 17.8 & 51.6 & 6.7& - & 39.2 & 33.4 \\
skt$\to$ & 46.4 & 11.1 & 40.9 & 15.0 & 58.2 & - & 34.3 & skt$\to$ & 54.7 & 15.2 & 44.5 & 15.1 & 61.4 & - & 38.2 & skt$\to$ & 53.0 & 14.8 & 45.5 & 15.6 & 62.7 & - & 38.3 \\
\rowcolor{blue!10}
Avg. & 35.2 & 10.7 & 34.2 & 8.0 & 49.4 & 29.6 & 27.8 & Avg. & 42.0 & 13.5 & 37.2 & 8.1 & 51.4 & 34.3 & 31.1 & Avg. & 41.2 & 13.0 & 38.4 & 8.9 & 53.6 & 32.9 & 31.3 \\
\midrule
\ours & clp & inf & pnt & qdr & rel & skt & Avg. & SHOT$^\dagger$$^\circ$ \cite{liang2020we} & clp & inf & pnt & qdr & rel & skt & Avg. & \ours$^\circ$ & clp & inf & pnt & qdr & rel & skt & Avg. \\
clp$\to$ & - & 15.2 & 45.5 & 14.2 & 65.9 & 43.2 & 36.8 & clp$\to$ & - & 15.5 & 41.7 & 16.9 & 61.8 & 41.7 & 35.5 & clp$\to$ & - & 15.8 & 46.0 & 17.2 & 66.3 & 44.7 & 38.0 \\
inf$\to$ & 41.3 & - & 45.1 & 3.9 & 62.5 & 33.7 & 37.3 & inf$\to$ & 41.8 & - & 38.4 & 4.5 & 58.6 & 32.4 & 35.1 & inf$\to$ & 42.9 & - & 44.7 & 4.3 & 63.1 & 35.7 & 38.2 \\
pnt$\to$ & 47.7 & 15.4 & - & 4.9 & 65.9 & 40.1 & 34.8 & pnt$\to$ & 49.4 & 16.4 & - & 7.4 & 62.1 & 39.4 & 34.9 & pnt$\to$ & 48.9 & 16.0 & - & 7.1 & 66.1 & 40.9 & 35.8 \\
qdr$\to$ & 22.6 & 1.1 & 7.7 & - & 17.4 & 15.3 & 12.8 & qdr$\to$ & 21.1 & 1.6 & 4.1 & - & 11.2 & 14.8 & 10.6 & qdr$\to$ & 21.0 & 1.0 & 5.6 & - & 18.2 & 16.2 & 12.4 \\
rel$\to$ & 54.8 & 17.4 & 52.0 & 6.7 & - & 41.0 & 34.4 & rel$\to$ & 51.4 & 18.5 & 47.1 & 8.4 & - & 38.6 & 32.8 & rel$\to$ & 55.4 & 17.7 & 52.3 & 8.7 & - & 41.9 & 35.2 \\
skt$\to$ & 55.3 & 15.1 & 46.1 & 16.2 & 65.4 & - & 39.6 & skt$\to$ & 54.3 & 14.7 & 43.3 & 17.9 & 60.1 & - & 38.1 & skt$\to$ & 55.9 & 15.1 & 46.2 & 19.1 & 64.9 & - & 40.2 \\
\rowcolor{blue!10}
Avg. & \best{44.3} & 12.8 & \best{39.3} & 9.2 & \best{55.4} & \best{34.7} & \best{32.6} & Avg. & 43.6 & \sest{13.3} & 34.9 & 11.0 & 50.8 & 33.4 & 31.2 & Avg. & \sest{44.8} & 13.1 & \sest{39.0} & \sest{11.3} & \sest{55.7} & \sest{35.9} & \sest{33.3} \\ 
\bottomrule
\end{tabular}}
\end{table*}

\noindent \textbf{c) Implementation details.} 
For the source model $f_s$, we train it using all samples from the source domain with a random seed of 1234 and select the checkpoint with the best performance on the source domain. 
In the case of DomainNet, the source model is trained on the training split and validated using the testing split.
Throughout this paper, we primarily use ResNet-50 \cite{he2016deep} as the backbone, as it is a widely adopted architecture in the UDA field.
Following \cite{liang2020we}, mini-batch SGD is employed to learn the layers initialized from the ImageNet pre-trained model or last stage with the learning rate (1e-3), and new layers from scratch with the learning rate (1e-2). 
Besides, we use the suggested training settings in \cite{long2018conditional,liang2020we}, including learning rate scheduler, momentum (0.9), weight decay (1e-3), bottleneck size (256), and batch size (64).
Concerning the parameters in \ours, we adopt the following values for all datasets: $r=1,\beta=0.5,\tau=0.1,\rho=0.5$.
Additionally, $T_m=30,\eta=0.95$ is used for all datasets, except for DomainNet, where $T_m=10,\eta=0.6$. 
We randomly run all the methods three times with different random seeds ({2024, 2025, 2026}) using \textbf{PyTorch} and report the average accuracies.

\subsection{Results on Standard UDA datasets}
We provide the results on three standard UDA datasets on Tables~\ref{tab:home},\ref{tab:office},\ref{tab:domainnet}. 
As shown in Table~\ref{tab:home}, \ours~achieves the best average accuracy under the non-hard-label scenario, outperforming the second-best method, SEAL, by approximately 0.5\%.
Methods utilizing strong data augmentations (i.e., BETA, SEAL, and \ours) clearly have an advantage over the other methods.
Without the use of strong data augmentations, ProD achieves the best performance, surpassing DINE.

In the hard-label scenario, we present results for only the well-performing methods and observe that \ours~consistently achieves the best average accuracy.
Although all methods experience a decline in accuracy when transitioning from non-hard-label to hard-label scenarios, it is noteworthy that the performance gap between \ours~and the second-best method enlarges.
Surprisingly, ProD significantly outperforms existing black-box counterparts such as BETA and SEAL.
This suggests that both \ours~and its distillation component, ProD, exhibit greater robustness to the quality of the source predictions.

\begin{table}[!ht]
\caption{Per-class accuracies (\%) on the \textbf{Office-Home-RSUT} dataset for UDA under two black-box scenarios.}
\label{tab:home-rsut}
\setlength{\tabcolsep}{2.0pt}
\resizebox{0.99\linewidth}{!}{
\begin{tabular}{lccccccca}
\toprule
Methods & Hard & Cl$\to$Pr & Cl$\to$Re & Pr$\to$Cl & Pr$\to$Re & Re$\to$Cl & Re$\to$Pr & Avg. \\
\midrule
No Adapt. & - & 51.5 & 51.4 & 37.1 & 67.2 & 39.6 & 71.2 & 53.0 \\
NLL-OT \cite{asano2019self} & \xmark & 53.3 & 55.4 & 38.9 & 68.8 & 40.5 & 71.8 & 54.8 \\
NLL-KL \cite{zhang2021unsupervised} & \xmark & 54.0 & 57.0 & 39.7 & 69.8 & 42.3 & 72.1 & 55.8 \\
NLL-MM \cite{liang2021source} & \xmark & 60.8 & 54.8 & 34.6 & 69.7 & 36.3 & 74.5 & 55.1 \\
SHOT$^\dagger$ \cite{liang2020we} & \xmark & 61.3 & 64.9 & \best{43.2} & 73.7 & \best{44.8} & 76.2 & 60.7 \\
DINE \cite{liang2022dine} & \xmark & 61.6 & 56.0 & 27.5 & 69.3 & 35.3 & 75.9 & 54.3 \\
BETA \cite{yang2023divide} & \xmark & 63.5 & 59.9 & 33.7 & 71.2 & 39.8 & 77.6 & 57.6 \\
SEAL \cite{xia2024separation} & \xmark & 62.8 & 62.1 & 39.3 & \best{74.3} & 44.1 & 77.1 & 59.9 \\
ProD & \xmark & 64.1 & 63.3 & 36.1 & 71.1 & 41.0 & 76.4 & 58.7 \\
\ours & \xmark & \best{65.2} & \best{66.3} & 39.5 & 73.2 & 44.4 & \best{78.1} & \best{61.1} \\
\midrule
SHOT$^\dagger$ \cite{liang2020we} & \cmark & 59.9 & 64.1 & 41.7 & 73.0 & 44.1 & 74.6 & 59.6 \\
DINE \cite{liang2022dine} & \cmark & 61.2 & 61.0 & 39.2 & 72.6 & 41.1 & 75.4 & 58.4 \\
BETA \cite{yang2023divide} & \cmark & 59.9 & 61.0 & 40.4 & 71.5 & 42.5 & 74.5 & 58.3 \\
SEAL \cite{xia2024separation} & \cmark & 60.1 & 60.0 & 40.0 & 72.2 & 43.1 & 74.7 & 58.3 \\
ProD & \cmark & 63.5 & 63.4 & 39.0 & 72.1 & 41.9 & 75.9 & 59.3 \\
\ours & \cmark & \sest{65.0} & \sest{65.7} & \sest{42.6} & \sest{74.6} & \sest{45.2} & \sest{77.9} & \sest{61.8} \\
\bottomrule
\end{tabular}}
\end{table}

As shown in Table~\ref{tab:office}, the results on the \textbf{Office} dataset further demonstrate the effectiveness of \ours, with notable performance gains over existing methods.
In the non-hard-label scenario, \ours~outperforms the second-best method, BETA, achieving the highest average accuracy. 
Additionally, ProD surpasses both DINE and SEAL, further demonstrating the effectiveness of prototypical distillation.
In the hard-label scenario, the performance advantage of \ours~over other methods becomes even more obvious, surpassing its own performance in the non-hard-label scenario, particularly when the target domain is W.
On the DomainNet dataset, similar conclusions can be drawn: \ours~achieves the best performance under both scenarios, with the distillation component, ProD, also delivering competitive results compared to other methods.

\begin{table*}[!ht]
\caption{Accuracies (\%) on the \textbf{Office-Home-Partial} dataset for UDA under two black-box scenarios.}
\label{tab:home-partial}
\setlength{\tabcolsep}{2.0pt}
\resizebox{0.99\linewidth}{!}{
\begin{tabular}{lccccccccccccca}
\toprule
Methods & Hard & Ar$\to$Cl & Ar$\to$Pr & Ar$\to$Re & Cl$\to$Ar & Cl$\to$Pr & Cl$\to$Re & Pr$\to$Ar & Pr$\to$Cl & Pr$\to$Re & Re$\to$Ar & Re$\to$Cl & Re$\to$Pr & Avg. \\
\midrule
No Adapt. & - & 46.5 & 71.3 & 80.8 & 56.0 & 60.6 & 66.8 & 59.2 & 40.2 & 76.5 & 71.4 & 48.7 & 76.9 & 62.9 \\
NLL-OT \cite{asano2019self} & \xmark & 55.4 & 78.4 & 86.5 & 67.3 & 73.1 & 78.1 & 70.3 & 49.1 & 83.8 & 76.6 & 57.0 & 82.0 & 71.5 \\
NLL-KL \cite{zhang2021unsupervised} & \xmark & 50.0 & 68.9 & 73.8 & 58.7 & 62.8 & 67.5 & 61.8 & 45.4 & 74.9 & 68.0 & 50.2 & 71.4 & 62.8 \\
NLL-MM \cite{liang2021source} & \xmark & 50.4 & 82.1 & 87.2 & 64.8 & 73.0 & 80.6 & 67.1 & 45.2 & 85.0 & 75.4 & 52.2 & 83.6 & 70.5 \\
SHOT$^\dagger$ \cite{liang2020we} & \xmark & 53.9 & 76.3 & 84.4 & 69.6 & 66.4 & 76.1 & 66.2 & 47.5 & 82.2 & 78.6 & 56.4 & 82.0 & 70.0 \\
DINE \cite{liang2022dine} & \xmark & 59.8 & 84.7 & 89.5 & 68.1 & 79.4 & 81.5 & 71.6 & 54.2 & \best{87.8} & 77.8 & 62.2 & \best{86.7} & 75.3 \\
BETA \cite{yang2023divide} & \xmark & 59.3 & 83.8 & 90.3 & 74.1 & 76.6 & 81.6 & 72.3 & 56.0 & 85.8 & 79.9 & 63.8 & 86.0 & 75.8 \\
SEAL \cite{xia2024separation} & \xmark & 58.0 & 80.0 & 83.4 & 73.6 & 73.7 & 77.5 & 71.5 & 56.9 & 84.3 & \best{80.2} & 61.4 & 81.1 & 73.5 \\
ProD & \xmark & 60.6 & \best{85.2} & 89.3 & 71.3 & \best{76.5} & \best{84.9} & 70.8 & 56.9 & 85.9 & 77.5 & 62.6 & 85.7 & 75.6 \\
\ours & \xmark & \best{65.2} & 84.7 & \best{90.5} & \best{76.2} & 75.7 & 84.3 & \best{74.5} & \best{60.0} & 86.6 & 80.0 & \best{64.7} & 85.2 & \best{77.3} \\
\midrule
SHOT$^\dagger$ \cite{liang2020we} & \cmark & 51.2 & 72.0 & 79.3 & 62.1 & 64.2 & 71.6 & 63.4 & 47.6 & 77.8 & 72.9 & 53.2 & 77.5 & 66.0 \\
DINE \cite{liang2022dine} & \cmark & 52.0 & 74.0 & 79.6 & 66.0 & 64.8 & 70.4 & 65.8 & 45.0 & 77.5 & 75.0 & 54.5 & 79.4 & 67.0 \\
BETA \cite{yang2023divide} & \cmark & 51.5 & 75.6 & 83.4 & 62.5 & 67.3 & 73.7 & 64.3 & 45.2 & 80.5 & 74.4 & 53.8 & 80.5 & 67.7 \\
SEAL \cite{xia2024separation} & \cmark & 48.7 & 72.7 & 79.5 & 60.9 & 64.8 & 69.4 & 61.3 & 45.9 & 78.3 & 72.9 & 52.2 & 77.1 & 65.3 \\
ProD & \cmark & 59.5 & \sest{83.6} & \sest{90.3} & 73.7 & \sest{76.5} & \sest{85.5} & 72.8 & 57.0 & \sest{87.2} & 79.0 & 60.6 & \sest{84.0} & 75.8 \\
\ours & \cmark & \sest{62.6} & 81.6 & \sest{90.3} & \sest{77.6} & 74.6 & 83.5 & \sest{73.0} & \sest{60.3} & 87.1 & \sest{81.2} & \sest{62.8} & 83.2 & \sest{76.5} \\
\bottomrule
\end{tabular}}
\end{table*}

\begin{table*}[!ht]
\centering
\caption{Accuracies (\%) on eight representative UDA tasks across three datasets. Using `proto.' denotes that $\beta = 0.5$, whereas in other cases, $\beta = 1.0$.}
\label{tab:ablation}
\setlength{\tabcolsep}{4pt}
\resizebox{0.99\linewidth}{!}{
\begin{tabular}{ccccccccccccccca}
\toprule
\multicolumn{4}{c}{ProD in Eq.~(\ref{eq:prod})} & \multicolumn{3}{c}{Ding in Eq.~(\ref{eq:ding})} & \multicolumn{2}{c}{\textbf{Office}} & \multicolumn{2}{c}{\textbf{Office-Home}} & \multicolumn{4}{c}{\textbf{DomainNet}} & \\
$\mathcal{L}_{skd}$ & $\mathcal{L}_{mix}$ & $\mathcal{L}_{mi}$ & proto. & $\mathcal{L}_{fm}$ & $\mathcal{L}_{afm}$ & $\mathcal{L}_{mi}$ & A$\to$D & W$\to$A& Ar$\to$Cl & Pr$\to$Re& clp$\to$pnt & pnt$\to$rel & rel$\to$skt & skt$\to$clp & Avg. \\
\midrule
& & & & & & & 80.3 & 63.5 & 44.7 & 73.5 & 34.7 & 57.2 & 34.3 & 47.9 & 54.5 \\
\cmark & & & & & & & 82.1 & 65.1 & 45.5 & 74.8 & 37.0 & 60.1 & 36.0 & 49.1 & 56.2 \\
\cmark & \cmark & & & & & & 84.5 & 65.7 & 46.6 & 75.5 & 39.8 & 60.4 & 37.6 & 50.4 & 57.6 \\
\cmark & & \cmark & & & & & 85.7 & 67.4 & 47.6 & 76.3 & 40.2 & 62.2 & 38.7 & 52.5 & 58.8 \\
\cmark & \cmark & \cmark & & & & & 87.6 & 68.3 & 49.1 & 77.4 & 42.9 & 62.6 & 39.9 & 53.7 & 60.2 \\
\cmark & \cmark & \cmark & \cmark & & & & 93.4 & 74.0 & 52.1 & 81.2 & 43.6 & 63.8 & 39.5 & 53.4 & 62.6 \\
\midrule
\cmark & \cmark & \cmark & \cmark & \cmark & & & 93.8 & 76.3 & 54.2 & 79.5 & 43.8 & 62.8 & 38.3 & 53.7 & 62.8 \\
\cmark & \cmark & \cmark & \cmark & & \cmark & & 94.0 & 75.8 & 55.8 & 80.8 & 46.6 & 65.4 & 42.3 & 56.1 & 64.6 \\
\cmark & \cmark & \cmark & \cmark & & & \cmark & 94.4 & 75.9 & 54.4 & 81.9 & 39.9 & 58.8 & 34.5 & 51.8 & 61.4 \\
\cmark & \cmark & \cmark & \cmark & \cmark & & \cmark & 94.2 & 76.1 & 55.8 & 82.4 & 46.4 & 65.3 & 41.0 & 55.5 & 64.6 \\
\cmark & \cmark & \cmark & \cmark & & \cmark & \cmark & 94.8 & 76.3 & 56.0 & 82.5 & 46.5 & 65.9 & 41.7 & 56.4 & 65.0 \\
\bottomrule
\end{tabular}}
\end{table*}

\subsection{Results on UDA datasets with Label Shifts}
As mentioned above, we also study the effectiveness of different black-box UDA methods for label shifts.
As shown in Table~\ref{tab:home-rsut}, SHOT$^\dagger$, which was a weak baseline in previous tables, achieves better performance than both BETA and SEAL.
\ours~still achieves the best performance under the non-hard-label scenario, winning 3 out of 6 tasks.
In the challenging hard-label scenario, different methods exhibit varying behaviors. DINE and BETA show slight improvements, while SEAL performs worse.
\ours~again achieves the best performance under the hard-label scenario, winning all 6 tasks.
This may be because the source predictions become more unpredictable, which perturbs the gap between these scenarios and affects the performance of the methods.
Besides, we adopt per-class accuracy, following existing methods \cite{tan2020class, prabhu2021sentry}, which may highlight differences in performance across different runs.

Partial-set domain adaptation \cite{cao2018partial,liang2020balanced} can be considered a special case of label shift, where some classes are absent in the target domain. 
As shown in Table~\ref{tab:home-partial}, \ours~achieves the best average accuracy under both scenarios.
The performance of SHOT$^\dagger$ is relatively worse compared to other baseline methods. While other baseline methods (e.g., DINE, BETA, and SEAL) experience significant performance drops, our methods (i.e., ProD and \ours) remain more stable.

\begin{table*}[!ht]
\caption{Accuracies (\%) on the \textbf{Office} and \textbf{Office-Home} dataset for UDA under two black-box scenarios (ResNet-34 used as source model).}
\label{tab:res34}
\setlength{\tabcolsep}{1.5pt}
\resizebox{0.99\linewidth}{!}{
\begin{tabular}{lcccccacccccccccccca}
\toprule
Methods & Hard & A$\to$D & A$\to$W & D$\to$A & W$\to$A & Avg. & Ar$\to$Cl & Ar$\to$Pr & Ar$\to$Re & Cl$\to$Ar & Cl$\to$Pr & Cl$\to$Re & Pr$\to$Ar & Pr$\to$Cl & Pr$\to$Re & Re$\to$Ar & Re$\to$Cl & Re$\to$Pr & Avg. \\
\midrule
No Adapt. & - & 68.9 & 70.4 & 47.2 & 51.7 & 59.5 & 41.7 & 59.2 & 68.7 & 45.0 & 56.9 & 60.3 & 45.7 & 37.1 & 68.35 & 60.3 & 44.4 & 75.1 & 55.2 \\
NLL-OT \cite{asano2019self} & \xmark & 79.3 & 77.4 & 51.0 & 55.9 & 65.9 & 43.9 & 61.8 & 69.8 & 48.3 & 59.9 & 63.3 & 47.9 & 39.5 & 70.0 & 61.5 & 46.9 & 76.0 & 57.4 \\
NLL-KL \cite{zhang2023black} & \xmark & 80.7 & 78.2 & 53.1 & 57.4 & 67.3 & 45.1 & 62.5 & 70.2 & 48.7 & 60.7 & 64.1 & 47.8 & 40.5 & 70.4 & 62.0 & 48.3 & 76.0 & 58.0 \\
NLL-MM \cite{liang2021source} & \xmark & 78.0 & 78.1 & 61.2 & 63.7 & 70.2 & 46.6 & 70.5 & 75.9 & 57.8 & 70.2 & 74.0 & 57.7 & 42.2 & 77.2 & 67.1 & 44.8 & 82.5 & 63.9 \\
SHOT$^\dagger$ \cite{liang2020we} & \xmark & 86.2 & 85.0 & 71.2 & 73.1 & 78.9 & 52.2 & 75.2 & 79.2 & 63.9 & 75.1 & 78.0 & 61.2 & 51.0 & 80.6 & 71.1 & 55.6 & 82.3 & 68.8 \\
DINE \cite{liang2022dine} & \xmark & 78.9 & 85.0 & 70.0 & 71.3 & 76.3 & 53.3 & 75.3 & 80.2 & 60.6 & 74.0 & 77.2 & 63.3 & 51.9 & 80.6 & 70.7 & 56.5 & 83.7 & 69.0 \\
BETA \cite{yang2023divide} & \xmark & 85.3 & 85.7 & 74.7 & 74.0 & 79.9 & \best{55.8} & 76.3 & 80.8 & 61.9 & 75.3 & 78.8 & 64.6 & 53.9 & 82.1 & 72.6 & 58.2 & \best{84.9} & 70.4 \\
SEAL \cite{xia2024separation} & \xmark & 86.3 & 83.2 & 72.8 & 73.9 & 79.0 & 55.2 & 76.7 & 81.2 & \best{65.5} & 75.5 & 79.5 & \best{65.8} & 54.3 & 82.1 & \best{73.3} & \best{59.0} & \best{84.9} & 71.1 \\
ProD & \xmark & 89.3 & 85.7 & 72.2 & 73.1 & 80.1 & 52.7 & 77.0 & 79.6 & 64.1 & 76.9 & 78.6 & 64.1 & 52.6 & 81.3 & 70.2 & 55.5 & 83.1 & 69.6 \\
\ours & \xmark & \best{90.9} & \best{86.8} & \best{75.5} & \best{74.8} & \best{82.0} & \best{55.8} & \best{77.5} & \best{81.9} & 65.1 & \best{78.7} & \best{79.7} & 65.7 & \best{54.7} & \best{82.6} & 70.5 & 58.0 & 84.3 & \best{71.2} \\
\midrule
SHOT$^\dagger$ \cite{liang2020we} & \cmark & 88.3 & 84.7 & 71.5 & 72.6 & 79.3 & 51.6 & 74.2 & 77.9 & 61.7 & 74.4 & 77.0 & 60.0 & 50.0 & 79.6 & 69.1 & 54.9 & 80.9 & 67.6 \\
DINE \cite{liang2022dine} & \cmark & 81.5 & 83.7 & 68.4 & 71.0 & 76.1 & 52.7 & 72.7 & 77.3 & 59.3 & 72.3 & 75.1 & 58.5 & 50.2 & 78.3 & 68.2 & 55.4 & 81.5 & 66.8 \\
BETA \cite{yang2023divide} & \cmark & 82.9 & 83.1 & 67.3 & 69.5 & 75.7 & 52.4 & 72.3 & 77.2 & 58.8 & 71.8 & 75.0 & 57.5 & 49.4 & 78.7 & 67.8 & 56.0 & 82.4 & 66.6 \\
SEAL \cite{xia2024separation} & \cmark & 81.2 & 80.2 & 65.6 & 68.0 & 73.7 & 49.4 & 70.1 & 75.5 & 57.1 & 69.6 & 72.9 & 57.8 & 47.9 & 77.4 & 66.8 & 53.9 & 81.0 & 65.0 \\
ProD & \cmark & 86.9 & 85.9 & 71.2 & 73.2 & 79.3 & 51.6 & 75.3 & 78.9 & 62.5 & 75.8 & 77.3 & 62.6 & 50.0 & 80.3 & 70.1 & 55.2 & 82.8 & 68.5 \\
\ours & \cmark & \sest{88.5} & \sest{88.6} & \sest{74.9} & \sest{76.0} & \sest{82.0} & \sest{55.9} & \sest{75.8} & \sest{81.1} & \sest{64.1} & \sest{78.7} & \sest{79.0} & \sest{64.7} & \sest{54.7} & \sest{81.9} & \sest{71.2} & \sest{58.3} & \sest{84.5} & \sest{70.8} \\
\bottomrule
\end{tabular}}
\end{table*}

\begin{figure*}[!htb]
\centering
\small
\setlength\tabcolsep{1mm}
\renewcommand\arraystretch{0.1}
\begin{tabular}{cccc}
\includegraphics[width=0.24\linewidth, clip]{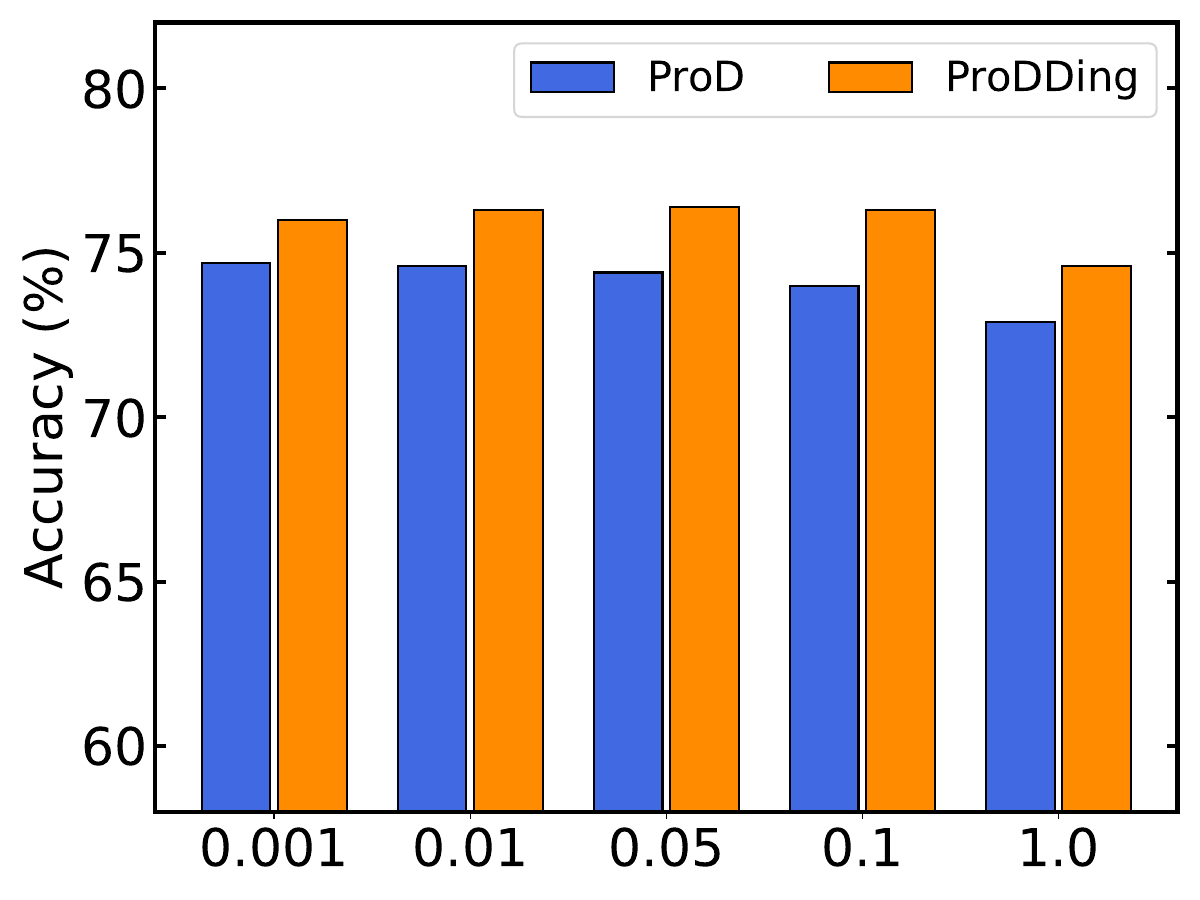} &
\includegraphics[width=0.24\linewidth, clip]{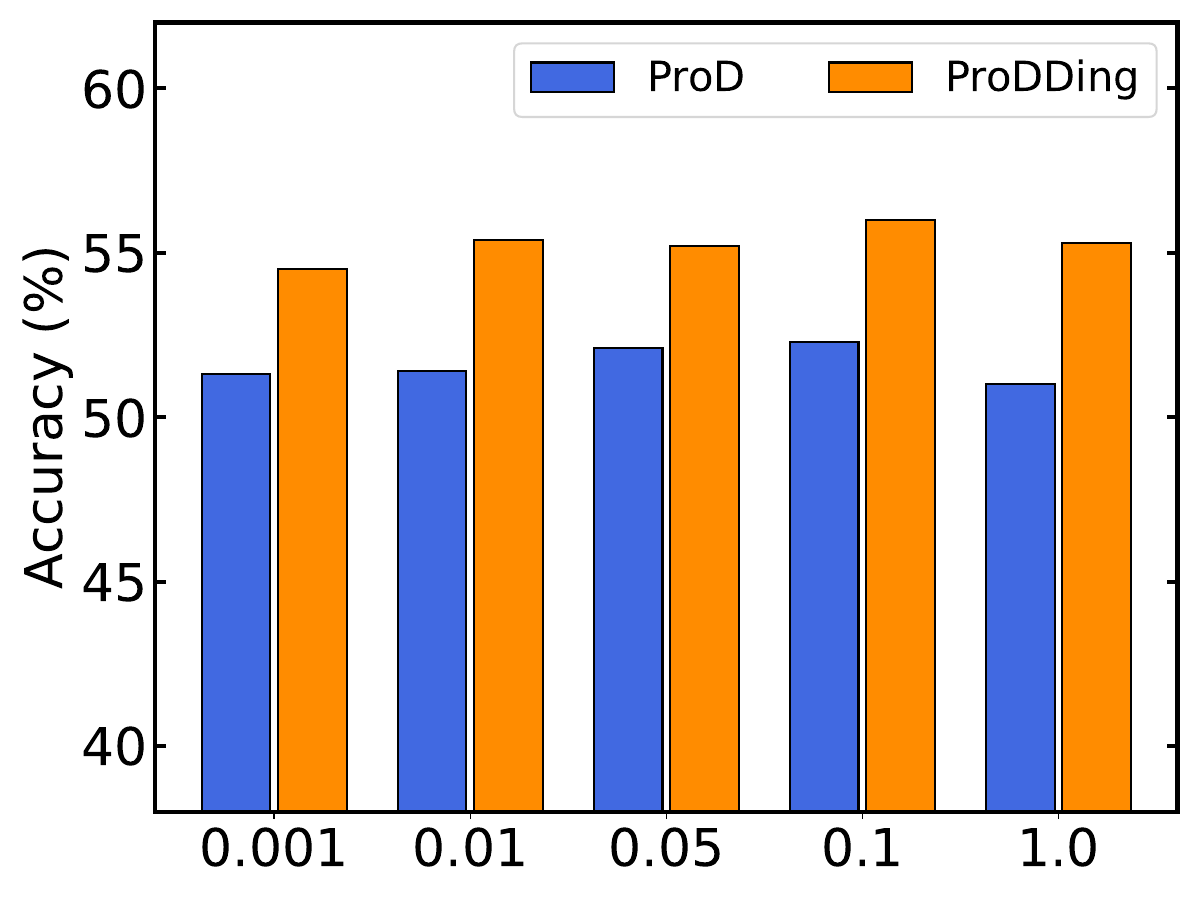} & 
\includegraphics[width=0.24\linewidth, clip]{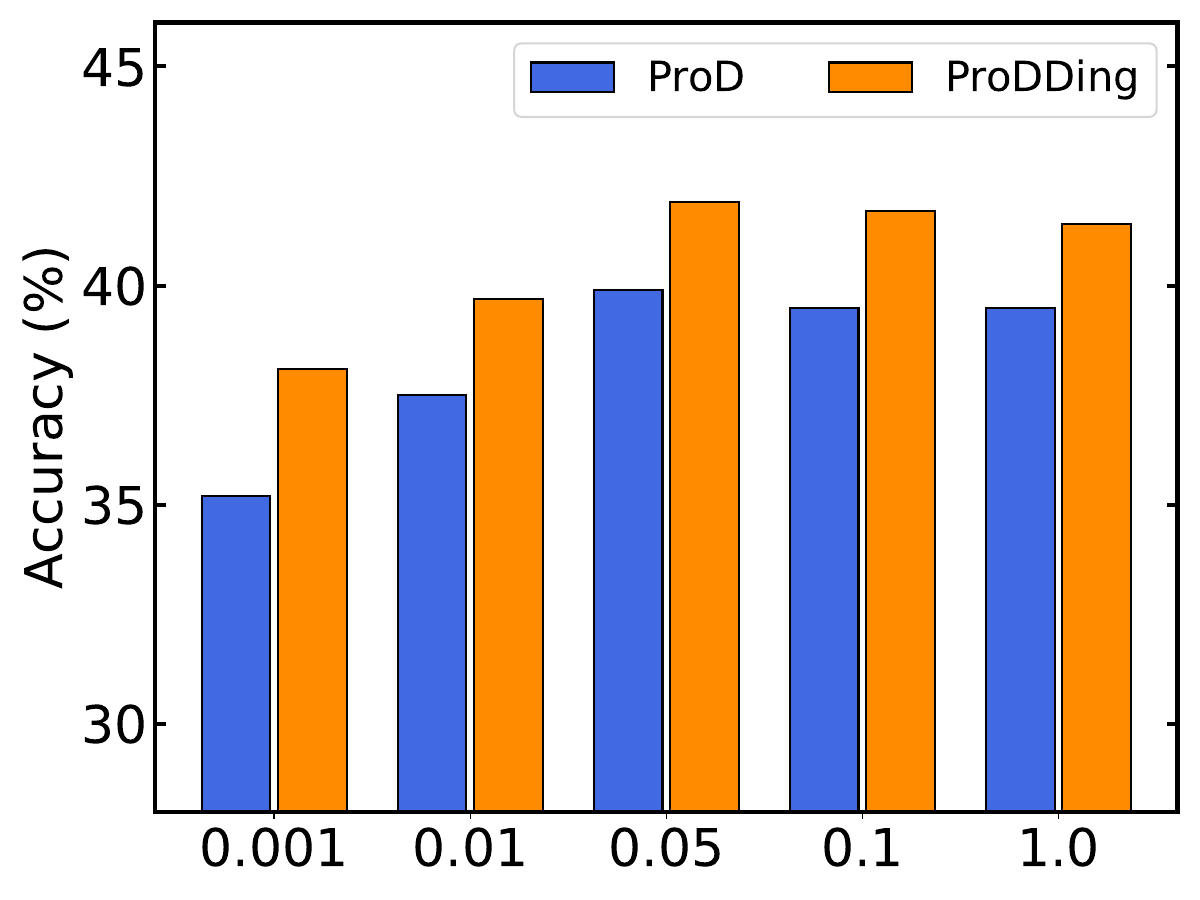} &
\includegraphics[width=0.24\linewidth, clip]{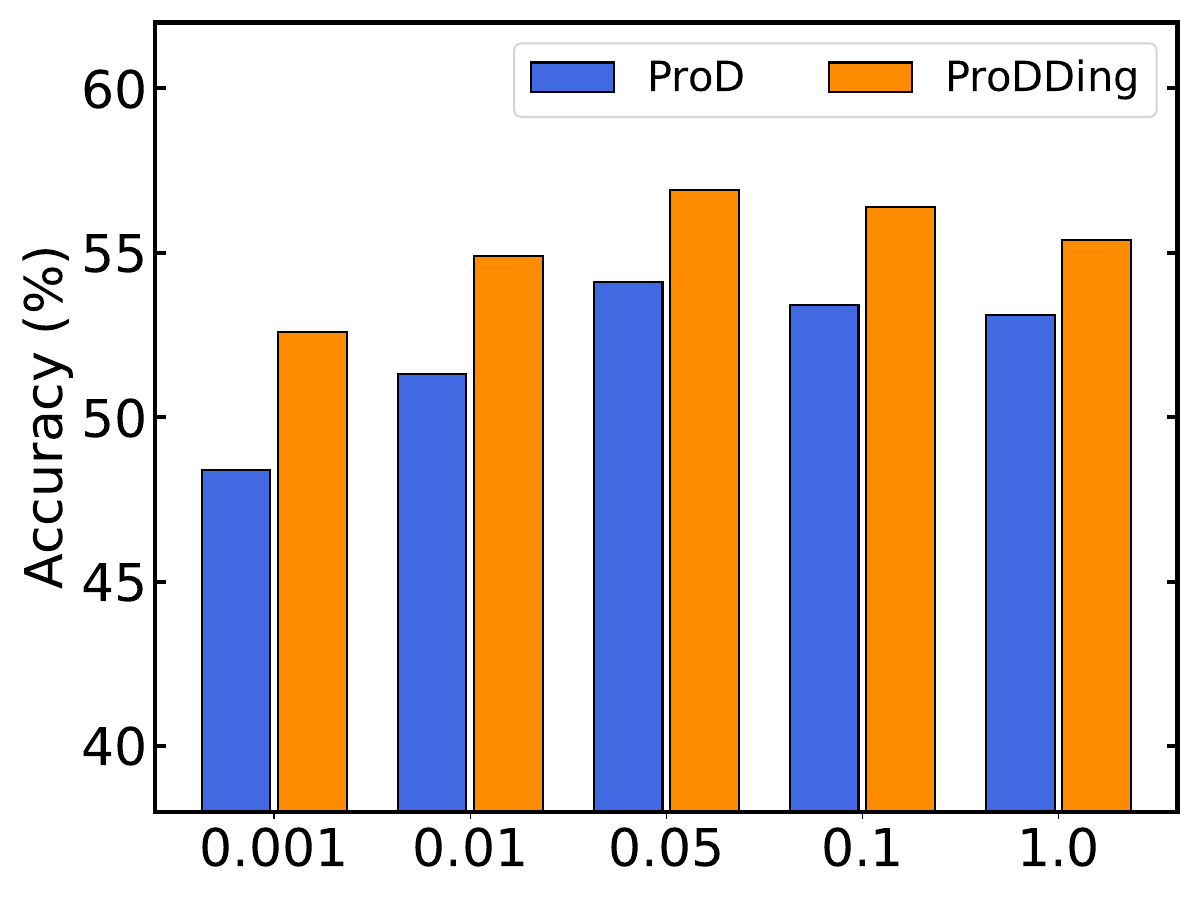} \\
~\\
(a) W$\to$A & (b) Ar$\to$Cl & (c) rel$\to$skt & (d) skt$\to$clp \\
\end{tabular}
\caption{Accuracies (\%) of ProD and \ours~under different temperature parameter $\tau$ for four representative UDA tasks across three datasets.}
\label{fig:sensitivity-T}
\end{figure*} 

\begin{figure*}[!htb]
\centering
\small
\setlength\tabcolsep{1mm}
\renewcommand\arraystretch{0.1}
\begin{tabular}{cccc}
\includegraphics[width=0.24\linewidth, clip]{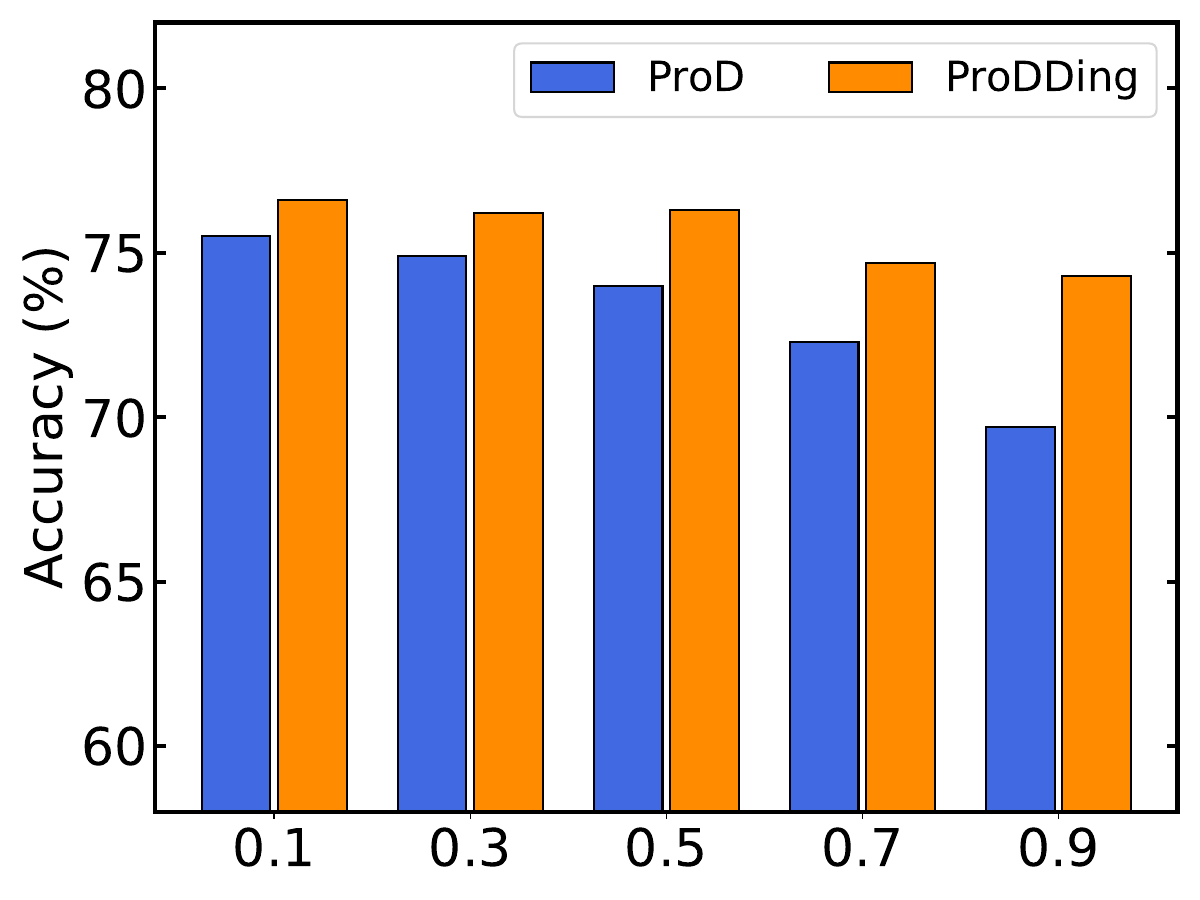} &
\includegraphics[width=0.24\linewidth, clip]{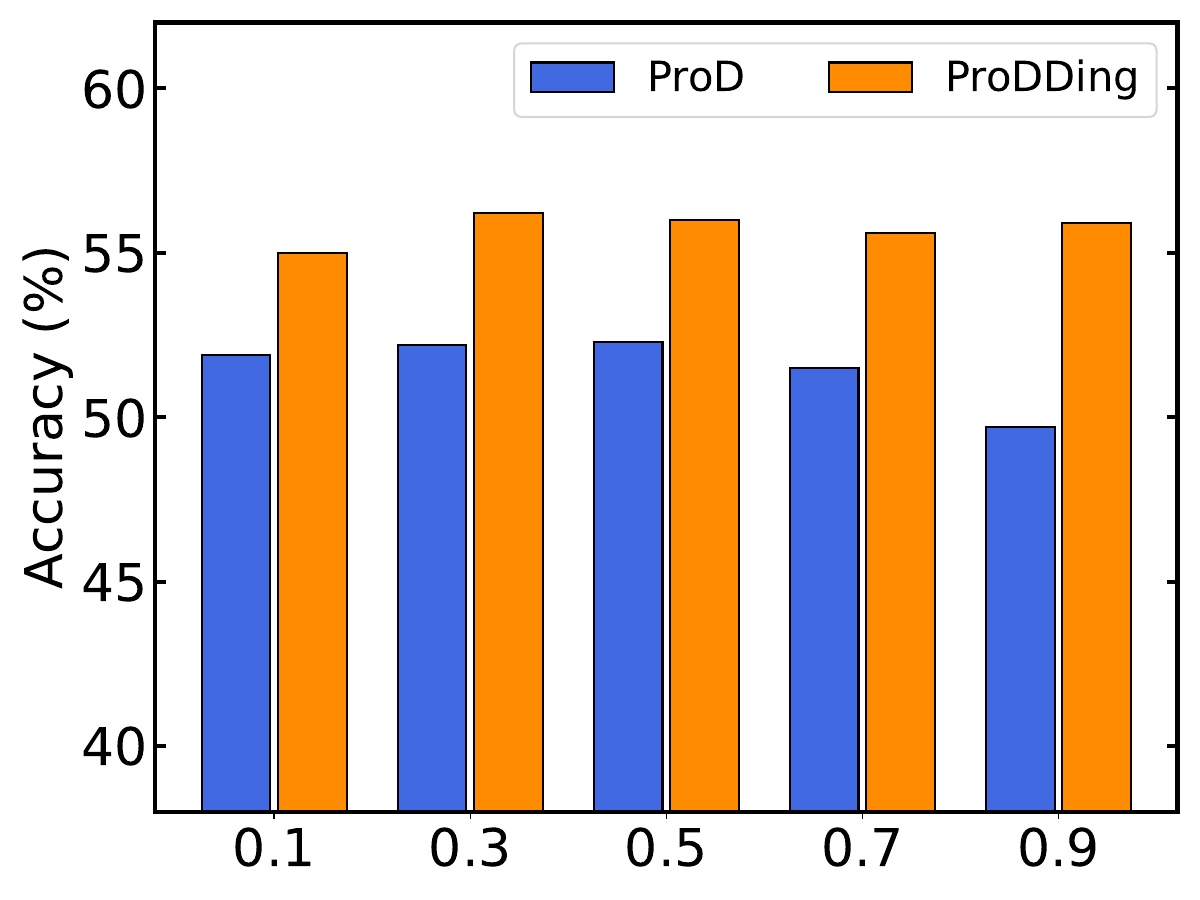} & 
\includegraphics[width=0.24\linewidth, clip]{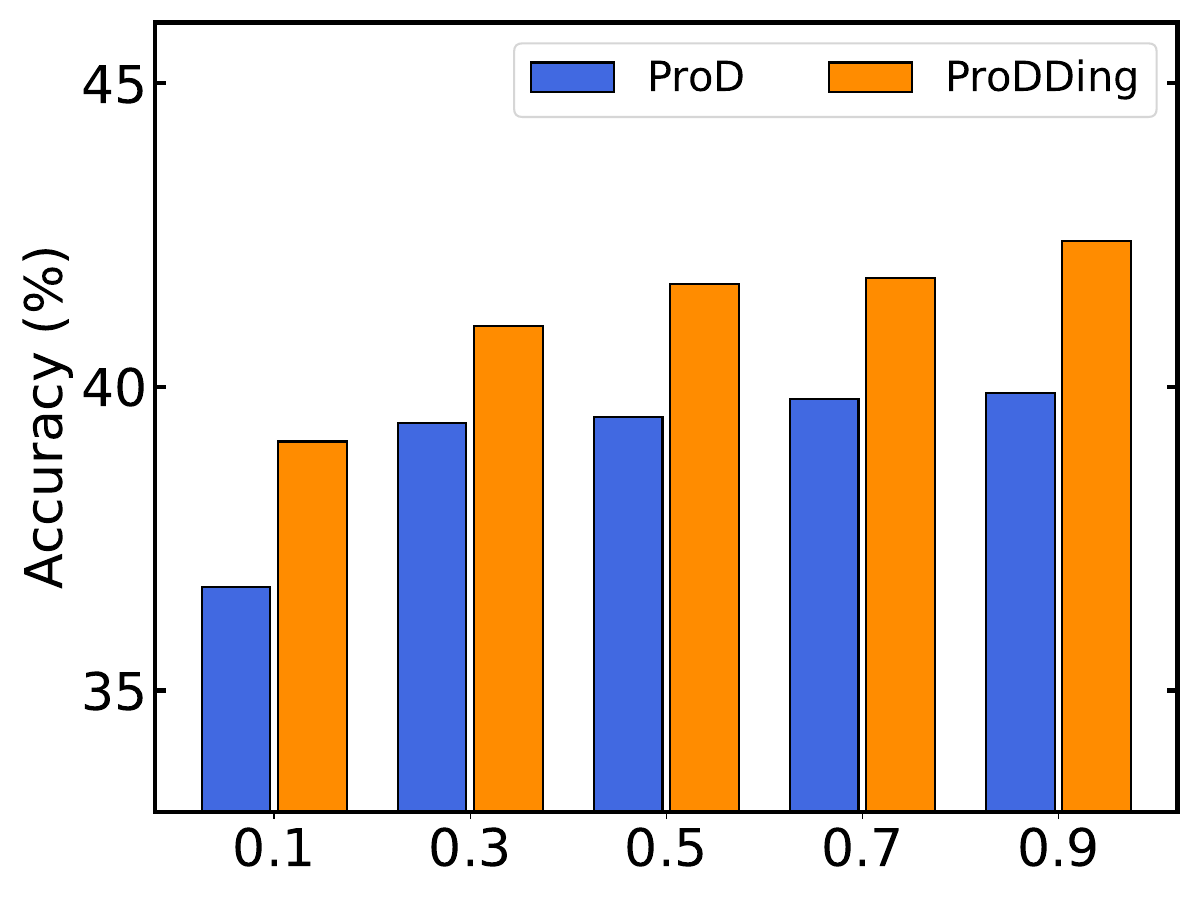} &
\includegraphics[width=0.24\linewidth, clip]{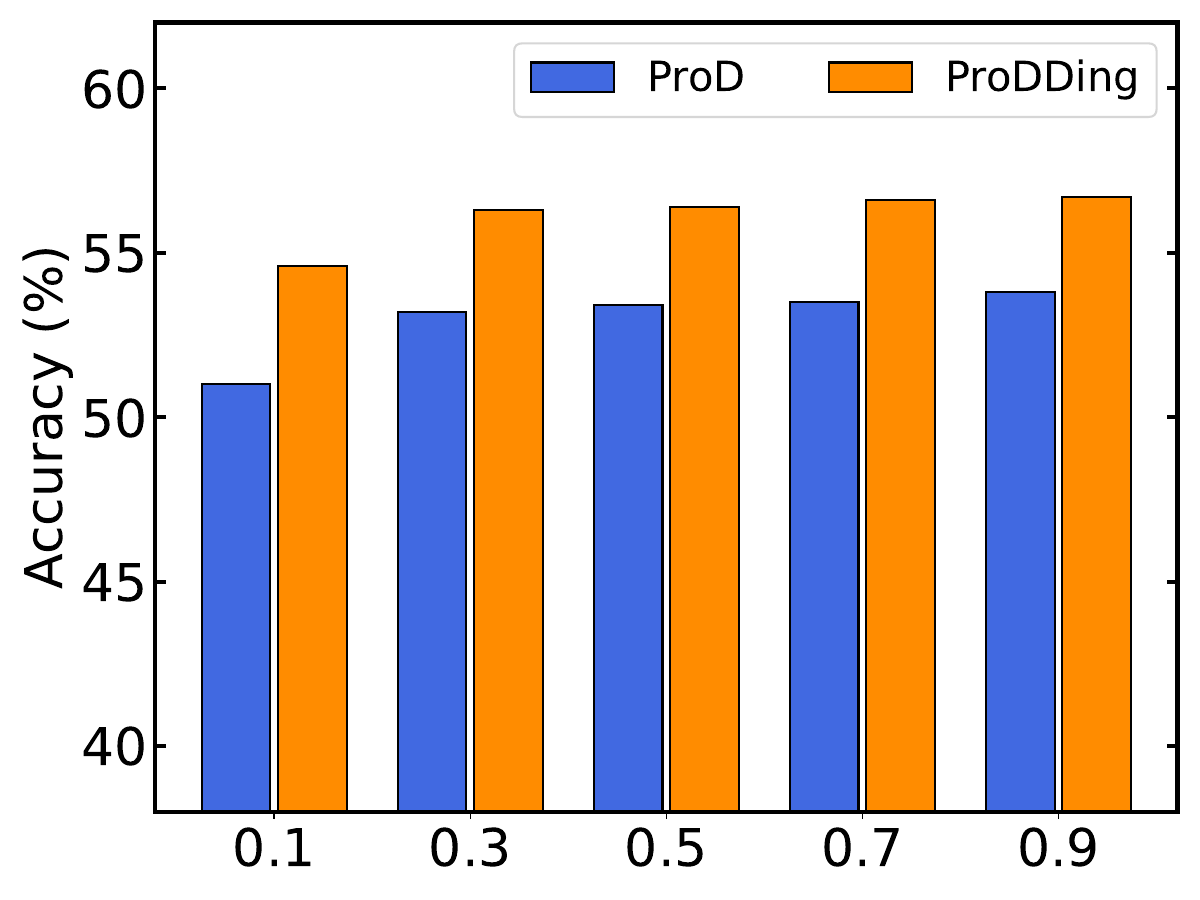} \\
~\\
(a) W$\to$A & (b) Ar$\to$Cl & (c) rel$\to$skt & (d) skt$\to$clp \\
\end{tabular}
\caption{Accuracies (\%) of ProD and \ours~under different balancing parameter $\beta$ for four representative UDA tasks across three datasets.}
\label{fig:sensitivity-init}
\end{figure*} 

\subsection{Ablation Study}
To validate the effectiveness of each component within the proposed \ours, we conduct extensive ablation experiments, as shown in Table~\ref{tab:ablation}.
The first line shows the results of `No Adapt.' 
With the introduction of the three objectives in Eq.~(\ref{eq:prod}), the average accuracy clearly increases.
The proposed prototypical pseudo-labeling in the initialized teacher also plays a crucial role, contributing to an improvement in accuracy by 2.4\%.
Regarding the second step, we find that using $\mathcal{L}_{afm}$ significantly outperforms $\mathcal{L}_{fm}$, which highlights the effectiveness of logit adjustment during the weak-to-strong consistency phase.
A similar improvement is observed with the $\mathcal{L}_{mi}$ objective.
When both debiased terms are combined, we achieve the best result in terms of average accuracy across 8 UDA tasks.
DINE \cite{liang2022dine} merely utilizes the $\mathcal{L}_{mi}$ in the fine-tuning step, but we find that incorporating the adjusted consistency term significantly helps boost performance.

\subsection{Analysis}
To study the effectiveness of \ours, we conduct experiments under the hard-label scenario unless stated otherwise.

\noindent $\rhd$ \textbf{Source architecture.} We further adopt the ResNet-34 backbone as the source model and present the results on the Office and Office-Home datasets in Table~\ref{tab:res34}.
Our method, \ours, consistently outperforms all baseline methods or achieves competitive performance under both non-hard-label and hard-label scenarios.
It is worth noting that under the hard-label scenario, \ours~achieves an accuracy of 70.8\% in Office-Home, which is close to the non-hard-label performance and 3.2\% higher than the second best method, SHOT$^\dagger$.
These results suggest that \ours~remains effective even under challenging conditions, such as when only limited information is available from the source domain due to either a weak source model or the use of hard labels.

\begin{figure}[!htb]
\centering
\small
\setlength\tabcolsep{1mm}
\renewcommand\arraystretch{0.1}
\begin{tabular}{c}
 \includegraphics[width=0.9\linewidth, clip]{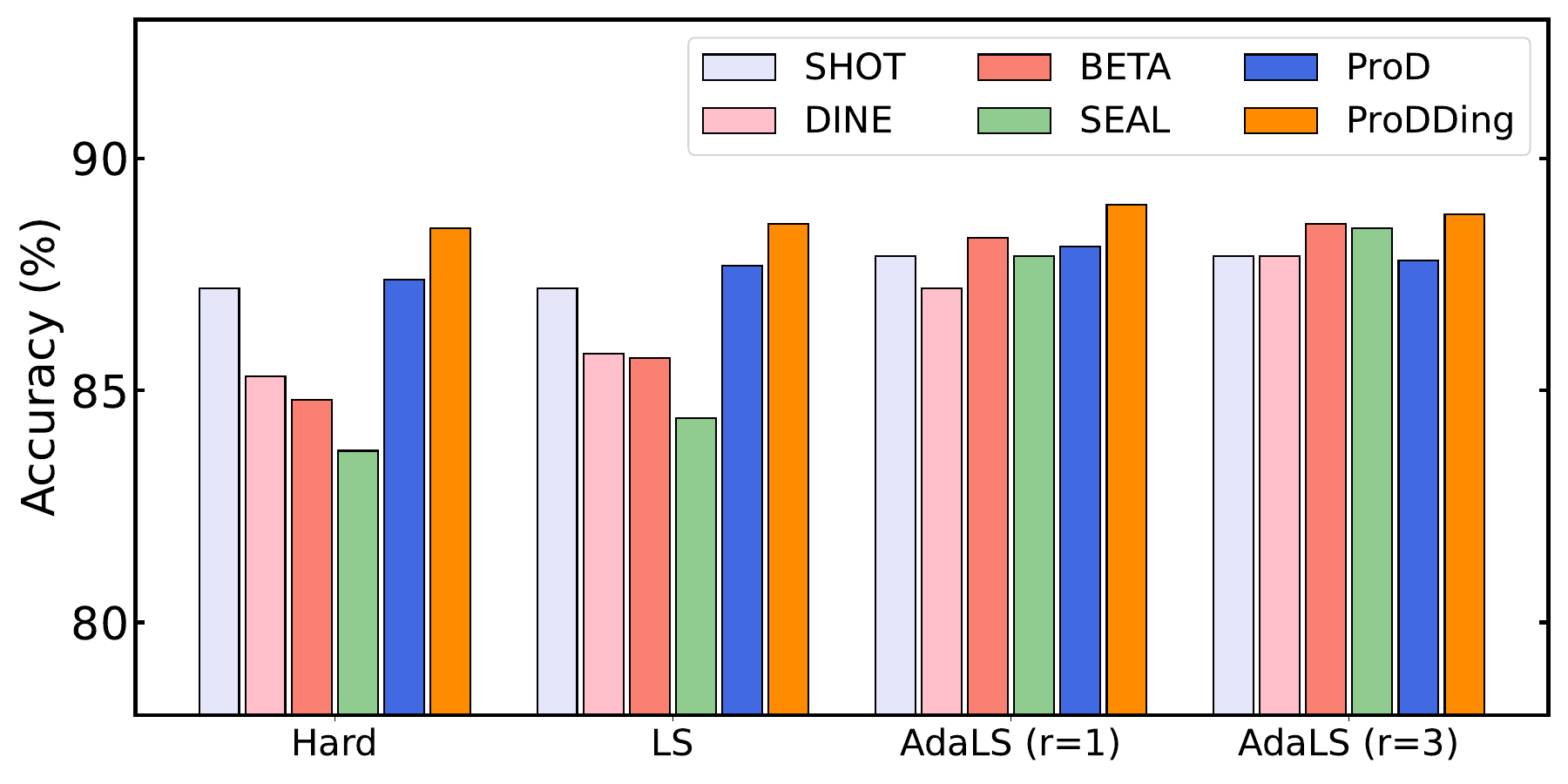} \\
 (a) \textbf{Office} \\
 \includegraphics[width=0.9\linewidth, clip]{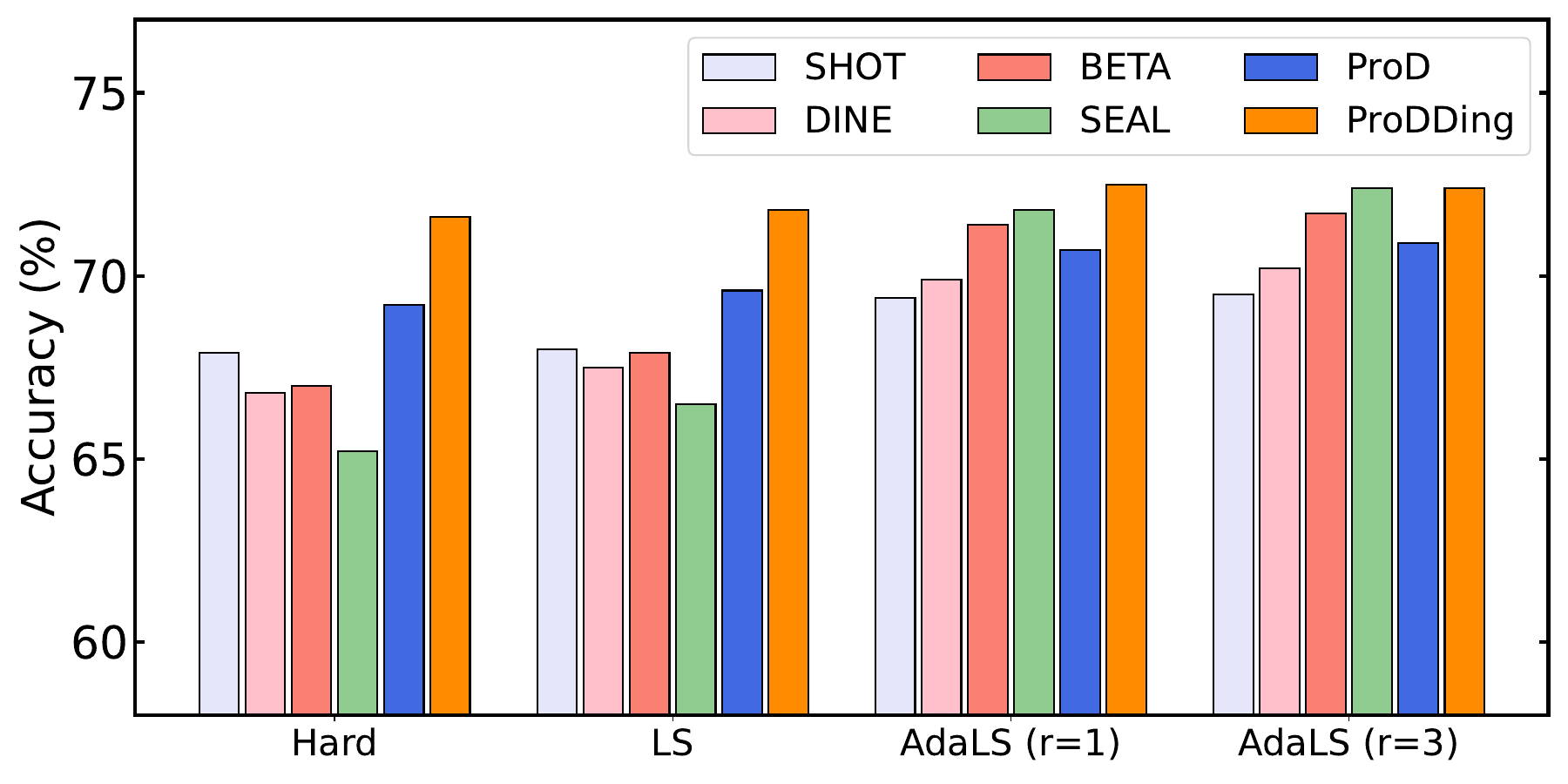} \\
 (b) \textbf{Office-Home} \\
\end{tabular}
\caption{Accuracies (\%) under different label smoothing techniques for \textbf{Office} and \textbf{Office-Home} datasets.}
\label{fig:label-smooth}
\end{figure} 

\noindent $\rhd$ \textbf{Sensitivity.}
We conduct a sensitivity analysis on the temperature parameter $\tau$ in Eq.~(\ref{eq:proto-pl}) and the balancing parameter $\beta$ in Eq.~(\ref{eq:init}).
Results across four adaptation tasks are shown in Fig.~\ref{fig:sensitivity-T} where $\tau$ is in the range of $[0.001, 0.01, 0.05, 0.1, 1.0]$, and Fig.~\ref{fig:sensitivity-init} where $\beta$ is in the range of $[0.1, 0.3, 0.5, 0.7, 0.9]$.
For each parameter setting, we provide the results for both ProD and \ours.
On the medium-sized datasets Office and Office-Home, the performance of our methods changes little across different temperature values.
Moreover, the results on the large-scale dataset DomainNet in Fig.~\ref{fig:sensitivity-T}(c-d) show that larger values of $\tau$ yield better results, with $\tau$=0.1 outperforming $\tau$=0.001 by approximately 4.7\% for ProD.
This can be attributed to the fact that a small $\tau$ leads to a sharper pseudo-prediction distribution, which results in overconfidence and a negative effect on the adaptation process.
Balancing parameter $\beta$ controls the initialization distribution ensembling ratio of the prediction obtained from the source model and through target domain feature clustering.
For tasks in the Office and Office-Home datasets, a smaller $\beta$ (favoring target domain clustering) performs better, while for the more challenging DomainNet dataset (Fig.\ref{fig:sensitivity-init}(c-d)), a larger $\beta$ (relying more on source domain predictions) yields superior results.
The uniform ensemble strikes a balance between these two strategies, ensuring \ours~adapts effectively to a wide range of scenarios.

\noindent $\rhd$ \textbf{Adaptive label smoothing.}
We investigate the impact of adaptive label smoothing and present a comparative analysis of \ours, alongside four baseline approaches.
We evaluate the performance of these methods under different smoothing techniques on two widely-used datasets, Office and Office-Home, as shown in Fig.~\ref{fig:label-smooth}.
Our results indicate that nearly all methods benefit from the confidence scores provided by the source model's interface. Specifically, on the Office-Home dataset, the accuracy of SEAL increases from 83.7\% (with hard labels) to 87.9\% (with AdaLS, $r$=1).
In contrast, \ours~achieves an improvement of 0.50\%.
This observation highlights a key limitation of baseline methods, such as SEAL, which rely heavily on the richness of the source domain information.
These methods struggle to maintain stable adaptation performance when the source model only provides hard labels.
Further analysis of vanilla label smoothing reveals that it outperforms the use of hard labels, confirming the positive impact of label smoothing on adaptation performance.
We also explore the scenario when the source model provides top-3 prediction with the confidence scores (AdaLS, $r$=3), which is also proved to provide a stable increase for almost all methods.
In conclusion, our study demonstrates the effectiveness of the label smoothing technique in enhancing adaptation performance.
Notably, \ours~outperforms all baseline methods and exhibits remarkable stability across different smoothing techniques, making it a robust solution for various scenarios.

\begin{figure}[!ht]
\centering
\small
\setlength\tabcolsep{1mm}
\renewcommand\arraystretch{0.1}
\begin{tabular}{c}
 \includegraphics[width=0.9\linewidth, clip]{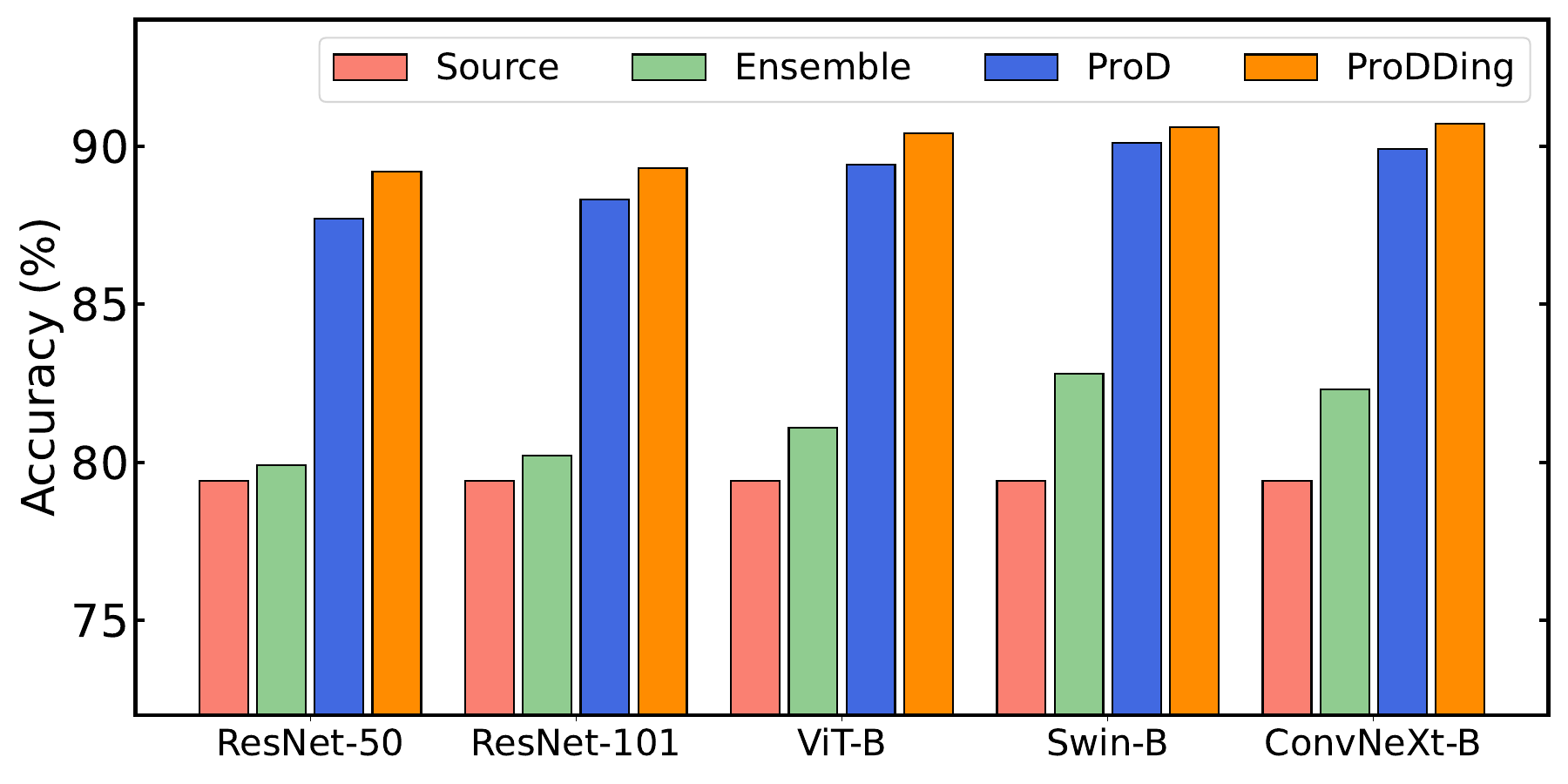} \\
 (a) \textbf{Office} \\
 \includegraphics[width=0.9\linewidth, clip]{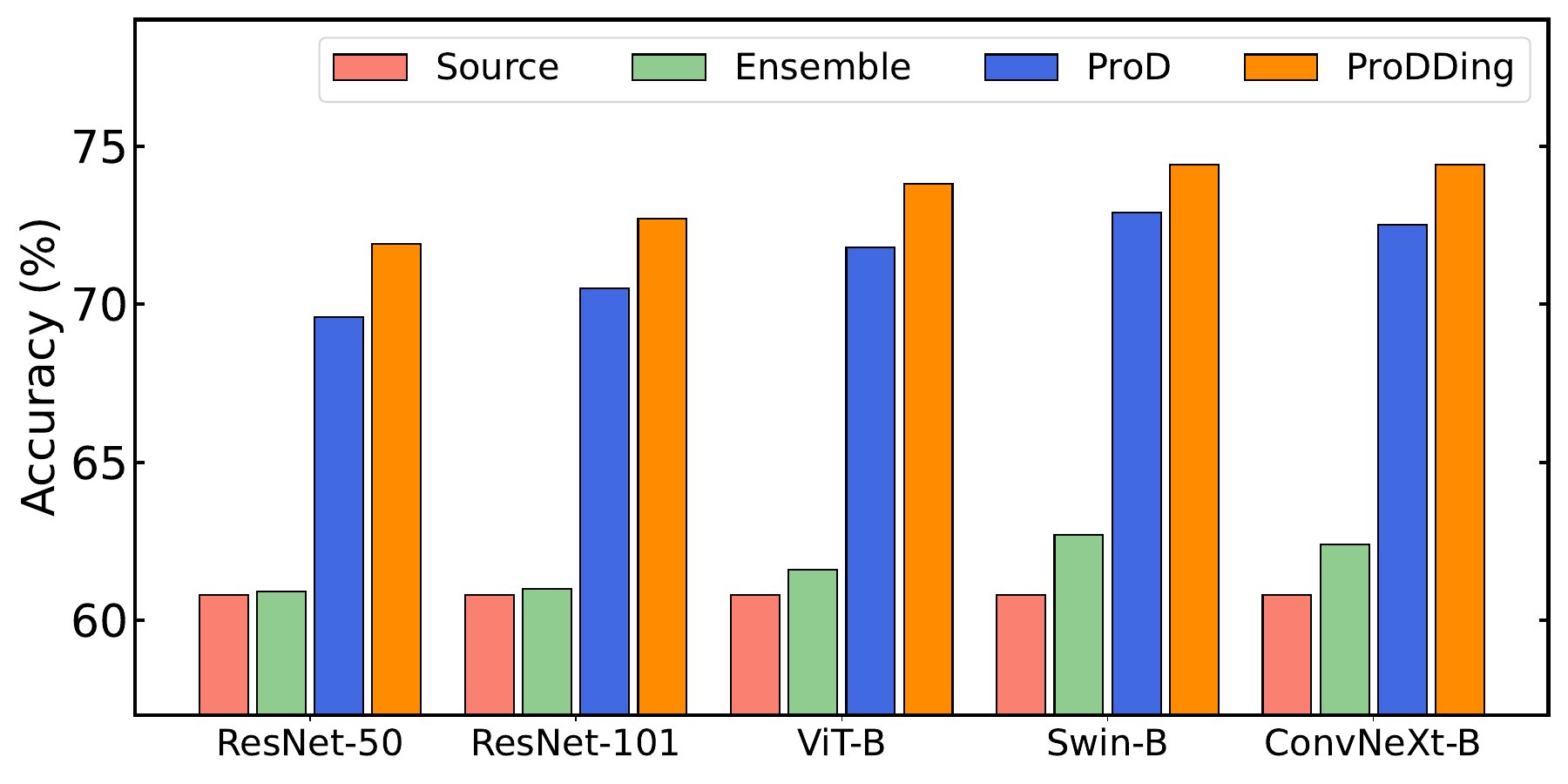} \\
 (b) \textbf{Office-Home} \\
\end{tabular}
\caption{Accuracies (\%) with different network architectures of feature extractor $g_t$ for \textbf{Office} and \textbf{Office-Home} datasets.}
\label{fig:encoder}
\end{figure}

\begin{figure}[!b]
\centering
\small
\setlength\tabcolsep{1mm}
\renewcommand\arraystretch{0.1}
\begin{tabular}{cc}
 \includegraphics[width=0.48\linewidth, clip]{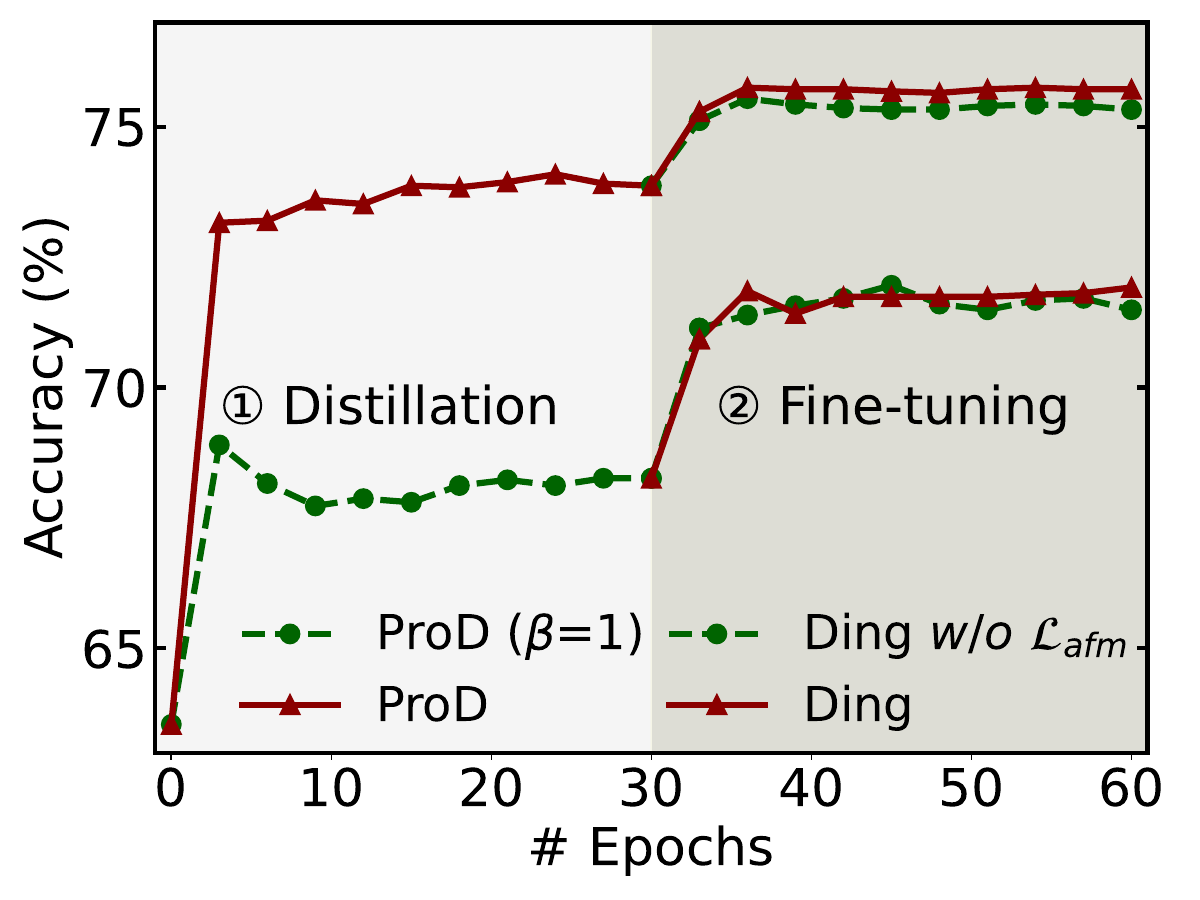} &
 \includegraphics[width=0.48\linewidth, clip]{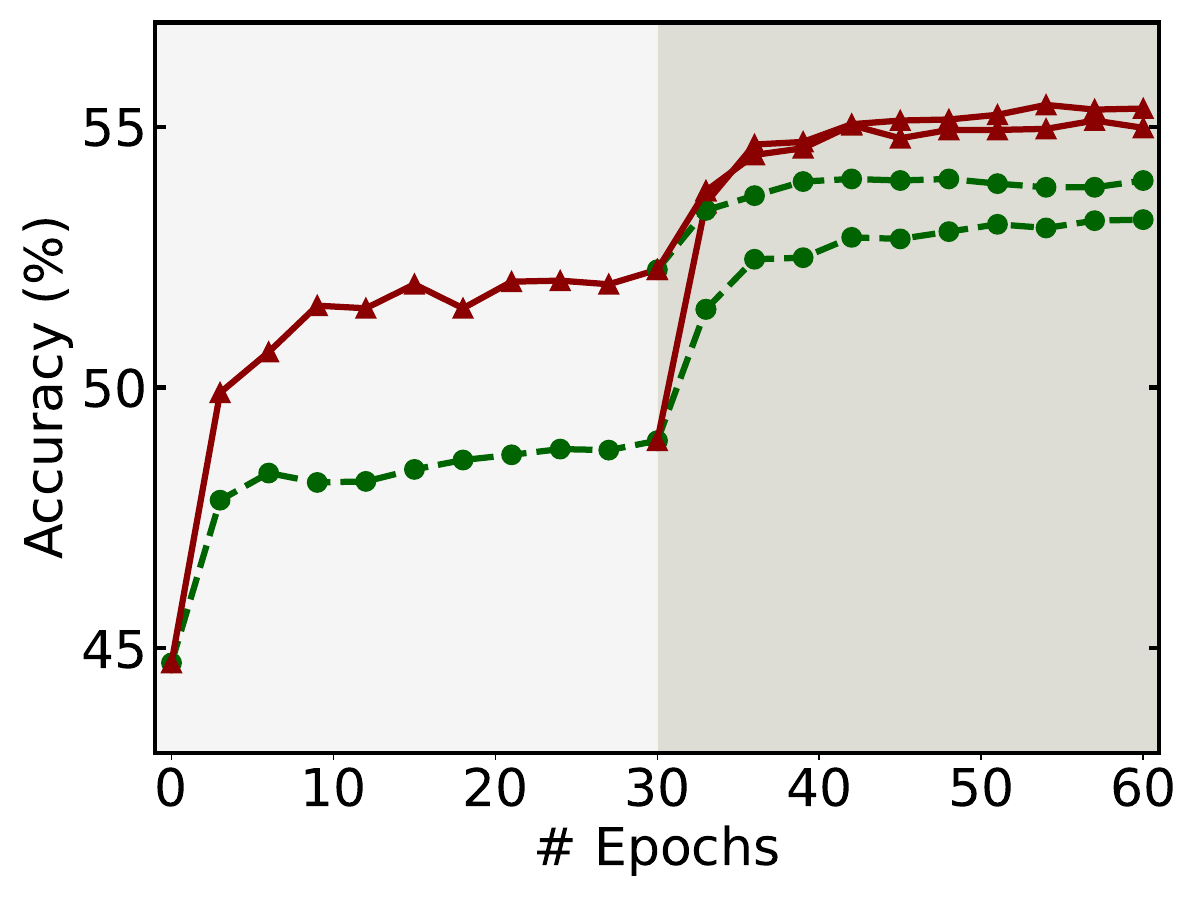} \\
 (a) W$\to$A & (b) Ar$\to$Cl \\
 \includegraphics[width=0.48\linewidth, clip]{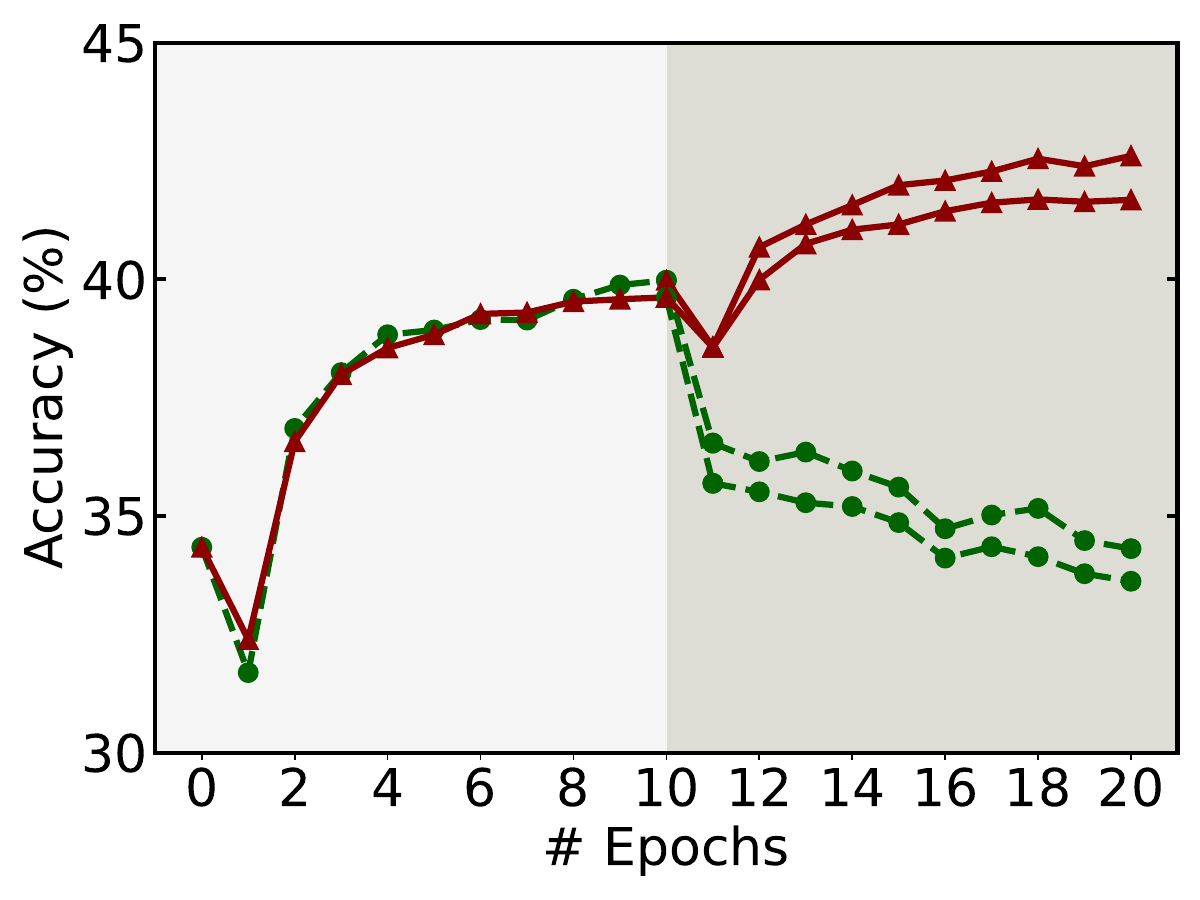} &
 \includegraphics[width=0.48\linewidth, clip]{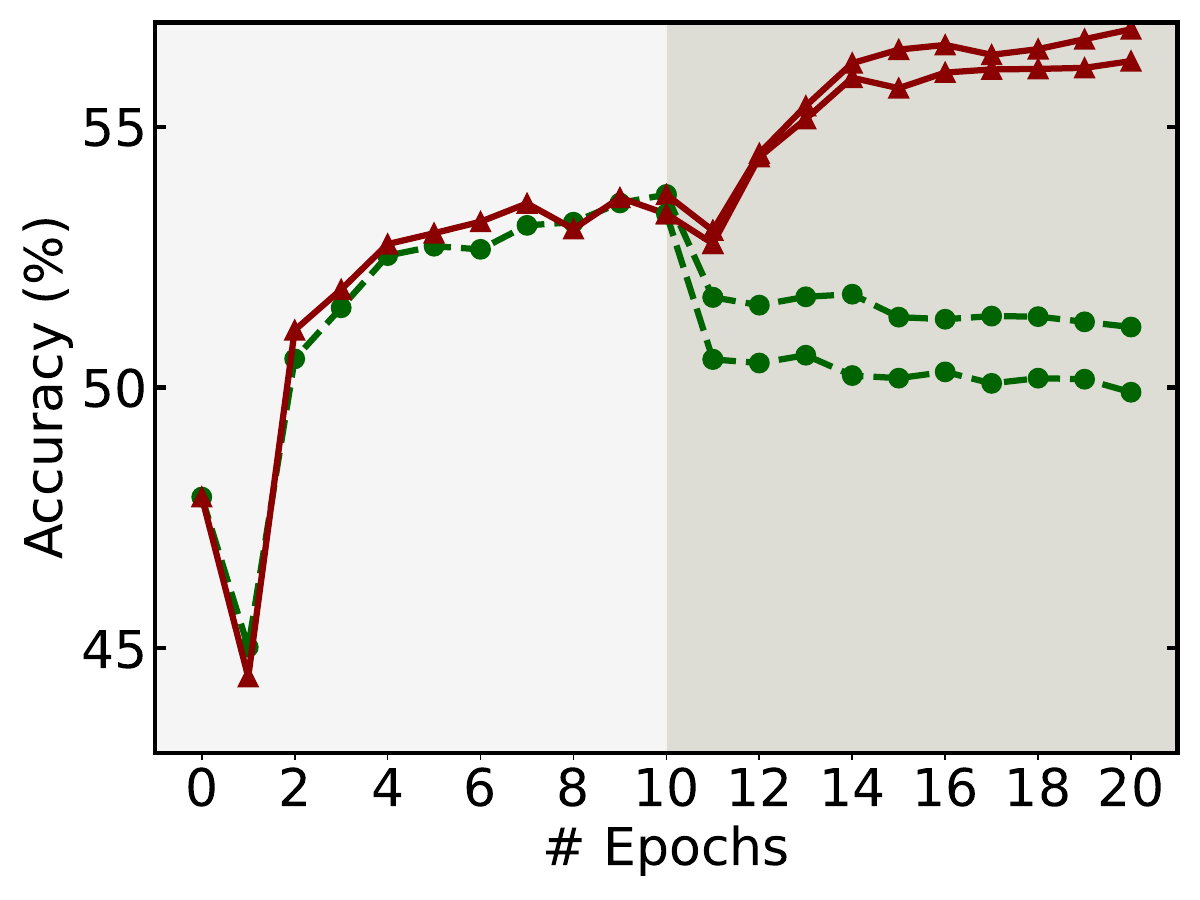} \\
 ~\\
 (c) rel$\to$skt & (d) skt$\to$clp \\
\end{tabular}
\caption{Accuracy convergence of DINE and \ours~for four representative UDA tasks across three datasets.}
\label{fig:acc_line}
\end{figure} 

\noindent $\rhd$ \textbf{Different pre-trained feature extractor architectures.}
To investigate the impact of various feature extractor architectures on the calculation of pseudo-predictions through feature clustering, we evaluate four additional network architectures: ResNet-101~\cite{he2016deep}, ViT-B~\cite{dosovitskiy2020image}, Swin-B~\cite{liu2021swin}, and ConvNeXt-B~\cite{liu2022convnet}.
The average results across two datasets, Office and Office-Home, are presented in Fig.~\ref{fig:encoder}.
Note that all experiments use ResNet-50 as the target model and query pre-trained feature extractors in a black-box manner.
It is shown that a stronger pre-trained feature encoder $g_t$ yields more accurate initial pseudo-predictions (Ensemble).
For example, on the Office-Home dataset, compared to ResNet-50, pseudo-prediction accuracy using clustering with the other four stronger feature extractors improves by 0.7\%, 1.9\%, 2.5\%, and 2.5\%, respectively.
This trend is also reflected in the accuracy of the final target predictions.
Our method achieves an accuracy of 73.8\% on Office-Home when using ViT-B as the feature extractor, which is an improvement over the 71.9\% achieved with ResNet-50.
This accuracy further increases to 74.4\% when combined with the Swin-B backbone, which benefits from a stronger ImageNet-1k classification ability.

\noindent $\rhd$ \textbf{Convergence.}
We provide the accuracy convergence curves for DINE \cite{liang2022dine} and \ours~in the distillation step and their accuracy curves based on two checkpoints in the fine-tuning step on four tasks.
In the distillation step, \ours~consistently outperforms DINE in the common tasks from Office and Office-Home and achieves competitive performance in challenging tasks from DomainNet.
Notably, \ours~achieves an accuracy of 73.9\% on W$\to$A task, which is 5.7\% higher than DINE.
As for the fine-tuning step, \ours~always improves accuracies and becomes convergent on all four tasks, while DINE suffers from negative transferring in complicated tasks from DomainNet, which indicates that it can not be deployed in challenging scenarios.
For example, in rel$\to$skt tasks, \ours~improves ProD from 39.6\% to 42.6\%, while DINE (Ding w/o $\mathcal{L}_{afm}$) drops to 33.6\% using the same checkpoint.

\section{Conclusion}
We explore a novel yet realistic UDA setting where the source vendor only provides its black-box predictor to the target domain, enabling the use of different networks for each domain while maintaining privacy.
Thereafter, we propose a simple yet effective two-step framework called \textbf{Pro}totypical \textbf{D}istillation and \textbf{D}ebiased tun\textbf{ing} (\ours).
Built on self-distillation, \ours~elegantly refines the noisy teacher output through adaptive smoothing and prototypical pseudo-labeling, while fully considering the data structure in the target domain during the distillation process.
To further mitigate potential class bias, \ours~continues fine-tuning the distilled model by penalizing logits that exhibit bias toward certain classes. 
Experiments across multiple datasets confirm the superiority of \ours~over existing approaches for various UDA tasks. 
Remarkably, even in the hard-label scenario, where only predicted labels are available, \ours~achieves surprisingly better results.

\bibliographystyle{IEEEtran}
\bibliography{tpami}

\end{document}